\documentclass[10pt,twocolumn,letterpaper]{article}

\usepackage[pagenumbers]{cvpr} 

\usepackage{graphicx}
\usepackage{amsmath}
\usepackage{amssymb}
\usepackage{booktabs}
\usepackage{multirow}
\usepackage{algorithm}
\usepackage{algpseudocode}

\usepackage[pagebackref,breaklinks,colorlinks]{hyperref}

\usepackage[capitalize]{cleveref}
\crefname{section}{Sec.}{Secs.}
\Crefname{section}{Section}{Sections}
\Crefname{table}{Table}{Tables}
\crefname{table}{Tab.}{Tabs.}

\def\cvprPaperID{*****} 
\def\confName{CVPR}
\def\confYear{2022}

\newcommand{\myparagraph}[1]{\vspace{0pt}\noindent{\bf #1}}

\DeclareMathOperator*{\argmax}{arg\,max}
\DeclareMathOperator*{\argmin}{arg\,min}
\DeclareMathOperator*{\softmax}{softmax}
\DeclareMathOperator*{\sigmoid}{sigmoid}
\DeclareMathOperator*{\Norm}{Norm}

\begin{document}

\title{TaleDiffusion: Multi-Character Story Generation with Dialogue Rendering}

\author{Ayan Banerjee$^1$, Josep Lladós$^1$, Umapada Pal$^2$, Anjan Dutta$^3$ \\
$^1$Computer Vision Center, Universitat Autònoma de Barcelona (\{abanerjee,josep\}@cvc.uab.es), \\ $^2$Indian Statistical Institute, Kolkata (umapada@isical.ac.in) \\ $^3$Institute for People Centred Artificial Intelligence, University of Surrey (anjan.dutta@surrey.ac.uk)\\}


\twocolumn[{%
\renewcommand\twocolumn[1][]{#1}%
\maketitle
\begin{center}
    \centering
    \captionsetup{type=figure}
    \includegraphics[width=\textwidth]{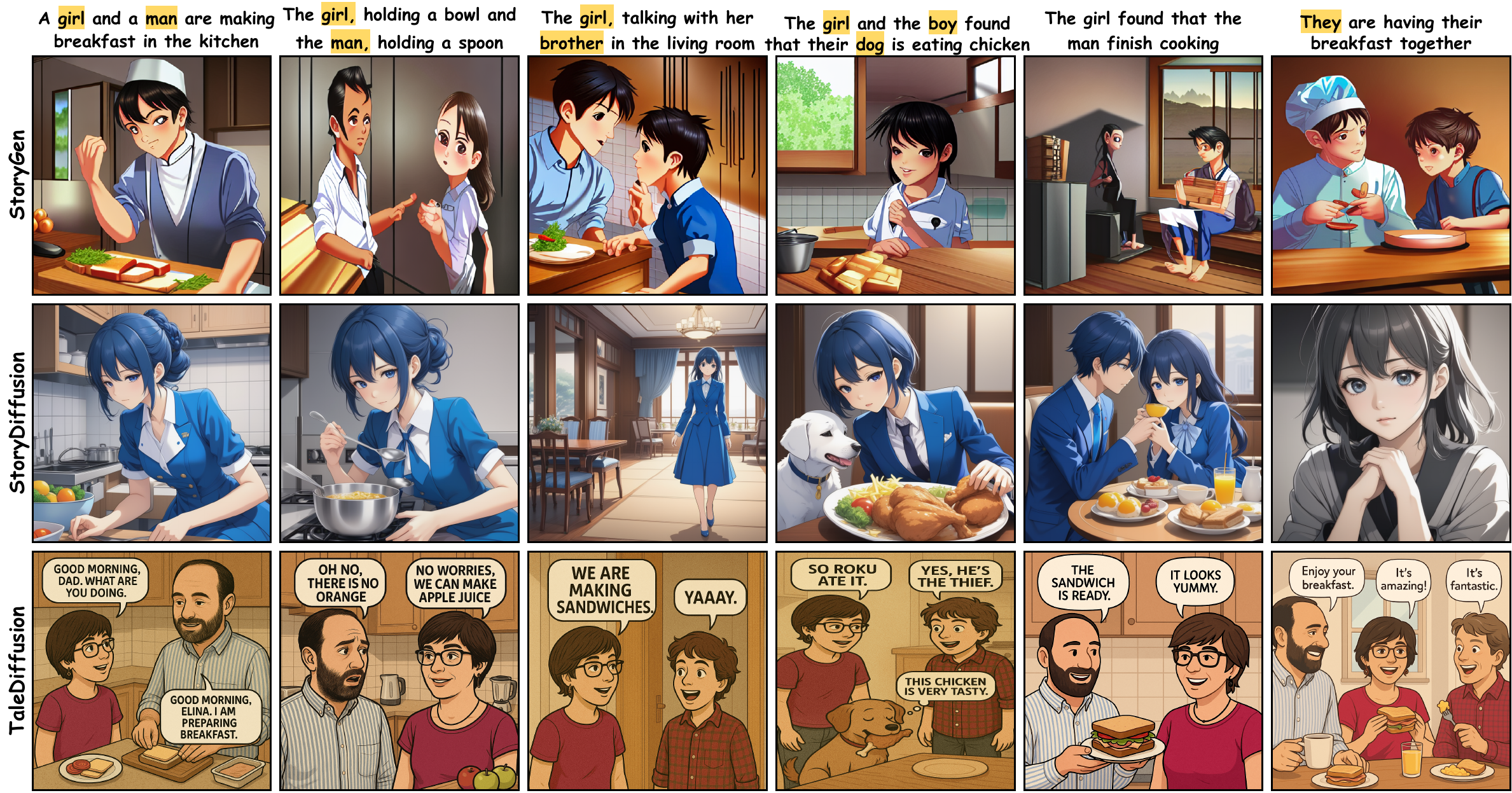}
    \captionof{figure}{TaleDiffusion enhances interactivity through dynamic character and environment handling using Identity Consistent Self-Attention, Region Aware Cross Attention, and multi-character customization via gradient fusion.}
    \label{fig:intro}
\end{center}%
}]

\begin{abstract}
Text-to-story visualization is challenging due to the need for consistent interaction among multiple characters across frames. Existing methods struggle with character consistency, leading to artifact generation and inaccurate dialogue rendering, which results in disjointed storytelling. In response, we introduce TaleDiffusion, a novel framework for generating multi-character stories with an iterative process, maintaining character consistency, and accurate dialogue assignment via postprocessing. Given a story, we use a pre-trained LLM to generate per-frame descriptions, character details, and dialogues via in-context learning, followed by a bounded attention-based per-box mask technique to control character interactions and minimize artifacts. We then apply an identity-consistent self-attention mechanism to ensure character consistency across frames and region-aware cross-attention for precise object placement. Dialogues are also rendered as bubbles and assigned to characters via CLIPSeg. Experimental results demonstrate that TaleDiffusion outperforms existing methods in consistency, noise reduction, and dialogue rendering. \href{https://github.com/ayanban011/TaleDiffusion}{github.com/ayanban011/TaleDiffusion.}
\end{abstract}

\section{Introduction}
\label{sec:intro}
Stories convey complex ideas, emotions, and narratives in an engaging and accessible way, making them valuable for education, entertainment, and communication~\cite{tatalovi2009scienceca,hosler2011comics}. Traditionally, story visualization requires significant human effort, involving collaboration between writers, illustrators, and editors. Recent advances in GANs~\cite{goodfellow2014gans} and diffusion models~\cite{rombach2022sd} offer promising avenues for automating this process~\cite{proven2021comicgan,zhou2024storydiffusion}, enabling dynamic, adaptive, and personalized storytelling with minimal manpower. Automated story generation can enhance education through engaging visuals, support rapid prototyping in creative industries, and enable interactive, personalized experiences in entertainment.

The current research landscape in story generation remains limited. GAN-based methods~\cite{proven2021comicgan} generate a single panel per story due to their mode collapse limitation, while video extraction techniques \cite{jing2015content,yang2021automatic} convert keyframes into panels with dialogues from audio, but neither can generate visual illustrations from written stories or text descriptions. A data-driven method \cite{cao2012automatic} generates stylistic manga layouts using existing manga pages as examples, though it only focuses on arranging the provided panels into a dynamic layout. Recent diffusion models for story generation have shown impressive capabilities in generating visually coherent sequences \cite{liu2024intelligent}, though they often introduce artifacts and inconsistent characters across frames \cite{wang2023autostory,cheng2024autostudio} during multi-character story generation. Also, diffusion-based dialogue rendering \cite{wu2024diffsensei} neither puts the accurate text in the bubble nor assigns the bubble correctly to the character, reducing smooth transitions across panels.

To address these limitations, we propose \textbf{TaleDiffusion} for realistic story visualization, which generates stories with multiple consistent and artifact-free characters across all frames, preserving spatial relationships between characters and providing accurate dialogue rendering with enhanced realism (see Fig. \ref{fig:intro}). We start by using a pretrained VLM to expand the input prompt, generate the layout of each frame, detailing character positions, scene semantics, and character-specific dialogues via iterative in-context learning (ICL). We introduce a per-box bounded attention-based mask latent generation technique for cleaner and more controlled outputs to control character interactions and reduce artifacts. We then introduce an identity-consistent self-attention (ICSA) mechanism to maintain character consistency across frames, ensuring that each character's appearance and attributes remain constant throughout the story, maintaining the stylization of the diffusion U-Net. For precise object placement within each scene, we propose a region-aware cross-attention (RACA) method that aligns objects with the scene context in each frame. Additionally, we propose a dialogue rendering mechanism that integrates dialogues in different languages into bubbles within the frames by enabling UTF-8 encoding. To accurately assign these bubbles to the corresponding characters, we use CLIPSeg \cite{luddecke2022image}, which ensures precise spatial alignment of the text within the visual narrative of the story.

In summary, we make the following contributions:
(1) We demonstrate the power of in-context learning over traditional chain-of-thought (CoT)~\cite{he2024dreamstory,cheng2024autostudio,singh2025storybooth} toward realistic multi-character story generation with appropriate dialogues. (2) We employ a per-box mask control mechanism combined with bounded attention over layout-based generation \cite{wu2024diffsensei,chen2024mangadiffusion} to eliminate common diffusion-based generation artifacts (extra limbs, distorted hands, and so on). (3) Propose the combination ICSA-RACA with subject-driven attention \cite{tewel2024training} to maintain character consistency between different frames and improve interactivity with accurate character placement. (4)  Conduct comprehensive quantitative and qualitative experiments, comparing our approach with recent benchmarks, and achieve SOTA performance.

\section{Related Work}
\label{sec:sota}

\myparagraph{Comic Generation:}
Early comic generation efforts began with ComicGAN \cite{proven2021comicgan}, which used adversarial networks for text-to-image generation. However, GAN’s mode collapse often resulted in single, low-quality panels. Video2Comics \cite{jing2015content,yang2021automatic} improved performance by converting keyframes into panels, adding dialogues from audio, and arranging them dynamically on a single page layout \cite{cao2012automatic}. Nonetheless, this approach lacked clarity in multi-character scenes and sometimes produced irrelevant text (\Eg, an entire page displaying only "The"; see  \cref{fig:vidcomp} in appendix). In contrast, our TaleDiffusion consistently generates multiple characters in a single panel without artifacts, utilizing mask guidance and identity-consistent self-attention, and accurately assigns rendered dialogues to characters via CLIPSeg \cite{luddecke2022image}.

\myparagraph{Story Visualization:}
It has progressed from early datasets like FlintstonesSV~\cite{gupta2018imagine} and PororoSV~\cite{li2019storygan} to models like StoryGAN~\cite{li2019storygan} and VLCStoryGAN~\cite{maharana2021integrating}, which improved text-image alignment but faced GAN-related issues. Transformer-based models like StoryDALL-E~\cite{maharana2022storydall} and diffusion models like StoryLDM~\cite{rahman2023make} enhanced frame continuity but struggled with character consistency. LLM-based methods~\cite{shen2023storygpt,pan2024synthesizing} improved coherence but remained limited when characters were unseen in training data. Recent approaches like StoryImager~\cite{tao2024storyimager} improved generation quality but were frame-limited. TaleDiffusion addresses these challenges by generating long stories with multiple customizable characters using ICSA and masked initialization. Unlike prior works~\cite{ruiz2023dreambooth,wei2023elite,tewel2024training,zhou2024storydiffusion}, it ensures consistency, reduces artifacts, and supports multi-character narratives via low-rank adaptation.

\begin{figure*}[!htbp]
    \centering
    \includegraphics[width=\textwidth]{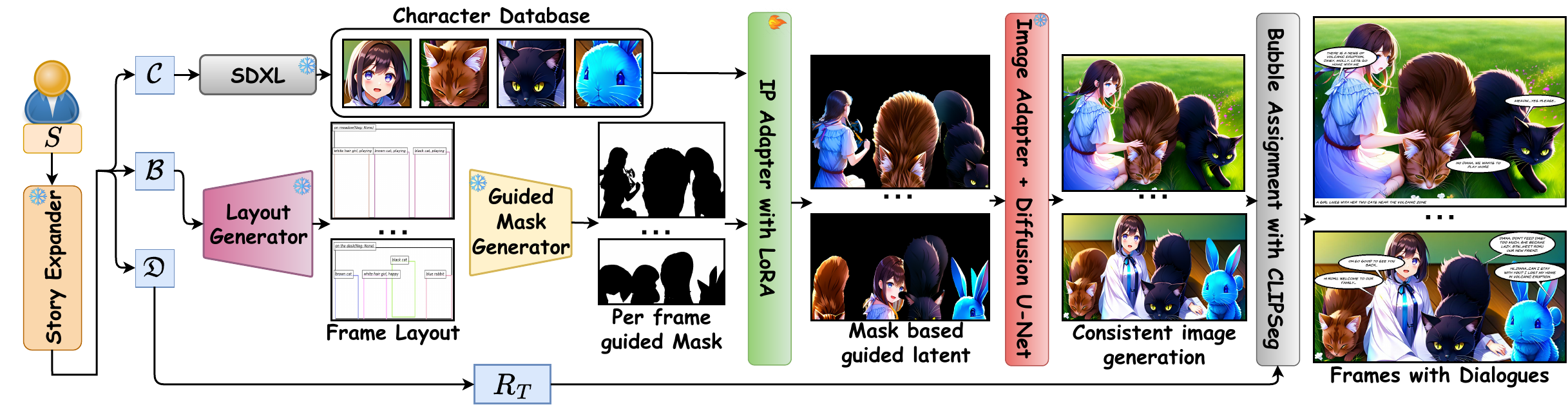}
    \caption{
    \textbf{TaleDiffusion framework:} Given a story $S$ and character descriptions $\mathcal{C}$, it uses a pretrained LLM to generate frame descriptions $\mathcal{B}$, dialogues $\mathfrak{D}$, and layouts $\Lambda_i$ with mask guidance. It builds an image character database from $\mathcal{C}$ via low-rank adaptation during latent guidance, which is processed by an image adapter and diffusion U-Net to denoise and add coherent backgrounds. Finally, it renders dialogue $R_T$, assigning bubbles to characters using CLIPSeg.}
    \label{fig:TaleDiffusion}
    \vspace{-4mm}
\end{figure*}

\myparagraph{Multi-Character Personalization:}
While single-character consistency has seen progress, multi-character consistency remains challenging. Li et al. \cite{li2024unbounded} simulate character lifecycles with Unbounded but lack frame-to-frame visual coherence. He et al. \cite{he2024improving} improve multi-subject consistency via Isolation and Reposition Attention, yet artifacts from interactions persist. TheaterGen \cite{cheng2024theatergen} generates narratives without visual grounding, while AutoStudio \cite{cheng2024autostudio} maintains character consistency but loses scene coherence. Storybooth \cite{singh2025storybooth} enables training-free consistency but lacks object placement control. Identity Decoupling \cite{jang2024identity} personalizes characters but is not optimized for interactive stories. DiffSensei \cite{wu2024diffsensei} and MangaDiffusion \cite{chen2024mangadiffusion} struggle with spatial consistency and dialogue alignment. Temporal methods like TokenFlow \cite{geyer2023tokenflow} and Animate Anyone \cite{hu2024animate} ensure continuity but neglect dialogue and narrative flow. In contrast, TaleDiffusion integrates an LLM for frame-wise description, bounded attention for interactions, ICSA for continuity, and CLIPSeg-based speech bubble rendering, achieving superior consistency and precise dialogue placement.

\section{TaleDiffusion}
\label{sec:method}
Given a story $S$, number of frames $n_p$, and a character set $\mathcal{C} = \{ C_1, \ldots, C_{n_c} \}$, TaleDiffusion uses a pretrained LLM \cite{wang2024large} to expand $S$ into frame descriptions $\mathcal{B} = \{ B_1, \ldots, B_{n_p} \}$ and dialogues $\mathfrak{D} = \{ \mathcal{D}_1, \ldots, \mathcal{D}{n_p} \}$, where $\mathcal{D}i = \{ D_{C_1, i_1}, \ldots, D_{C_{n_c}, i_j} \}$ contains dialogues in frame $i$, with $D_{C_k, i_j}$ being the $j$-th dialogue by $C_k$. Layouts $\Lambda_i$ for each $B_i$ are generated via iterative correction. Guided masks with bounded attention are used for each $\Lambda_i$ to reduce diffusion artifacts and support scalable generation (see \cref{fig:TaleDiffusion}). A character database $\mathcal{C}_\text{db}$ is built using a pretrained diffusion U-Net, aligning characters via LoRA and ICSA for latent guidance, further processed by an image adapter and diffusion U-Net. Finally, a dialogue rendering algorithm $\mathcal{R}_T$ places $D_{C_k, i_j}$ in the $i$-th frame and links it to $C_k \in \mathcal{C}$ using CLIPSeg \cite{luddecke2022image}. The end-to-end algorithm of the TaleDiffusion is depicted in \cref{algo4}.

\begin{algorithm}[!htbp]
\caption{TaleDiffusion Algorithm}
\label{algo4}
\begin{algorithmic}[1]
\Require  $S$, $n_p$,   $C = \{C_1, \dots, C_{nc}\}$ 
\Ensure Consistent frames with dialogues

\State Generate frame description and dialogues using ICL:
\State For each $B_i$, generate layout $\Lambda_i$:
\State Generate per-box masks  with bounded attention:
\State $A_{i,j} = \text{softmax}\left(\frac{Q_{i,j}^T K_{i,j}}{\sqrt{d}}\right) V_{i,j}$
\State $\quad M_{i,j} = \text{Norm}(\sigma(\xi \cdot \text{Norm}(A_{i,j} - \phi_j)))$
\State Generate latent \( Z_{\text{fg}} \) using ICSA as depicted in \cref{eq:4}:
\State Background denoising with \cref{eq:new}:
\State Render dialogue bubbles using CLIPSeg:
\State $(x, y), \Lambda = \text{CLIPSeg}(I, \{D_c\} \cup \{\text{"Face"}\})$ 
\State $\quad B_c = (x_b, y_b, W_{\text{bubble}}, H_{\text{bubble}}, (x_e, y_e), T_{\text{wrapped}}, (x, y))$
\State \textbf{Return:} Consistent frames with dialogues.
\end{algorithmic}
\end{algorithm}

\subsection{Story Expansion and Layout Generation}
\label{sec:m1}
\begin{figure*}[!htbp]
\centering
\includegraphics[width=\textwidth]{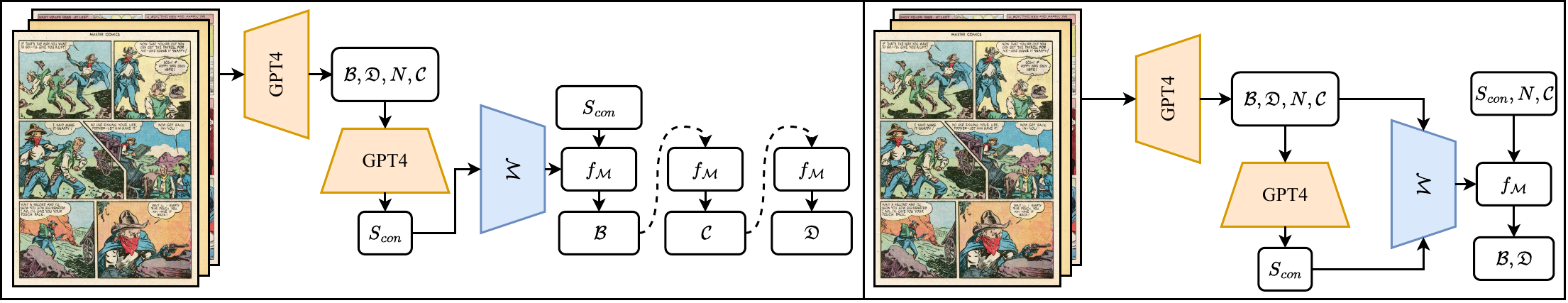}
\caption{\textbf{CoT (left) vs ICL (right):}  In CoT, the whole task is divided into multiple subtasks where each subtask is dependent on the previous one. So it can never have the full context. In contrast, ICL provides the whole context to the LLM in a single step and helps to provide more creativity and completeness.
}
\label{fig:LLMs}
\vspace{-4mm}
\end{figure*}

\myparagraph{Story Expansion:} Given a story $S$, characters $\mathcal{C}$, and frame count $n_p$, we generate detailed frame descriptions $\mathcal{B}$ and dialogues $\mathfrak{D}$. Unlike prior CoT-based methods \cite{zhang2023multimodal,wang2023autostory,he2024dreamstory,cheng2024autostudio} that lack coherence and creativity, we adopt in-context learning \cite{dong2024iclsurvey} (see \cref{fig:LLMs}) using the DCM772 dataset \cite{nguyen2018digital}. For each comic page, we extract frame descriptions, characters, and dialogues using an MLLM \cite{achiam2023gpt}, and summarize the story. These are used to construct demonstrations $\mathcal{X}_\text{con}$ for guiding generation. For a given input $(S, n_p, \mathcal{C})$, the likelihood of candidate outputs $(\mathcal{B}_j, \mathfrak{D}_j)$ is scored by a function $f_{\text{LLM}}$, and the highest-scoring output is selected:
\vspace{-2mm}
\begin{equation}
\vspace{-2mm}
    \mathcal{B}, \mathfrak{D} = \argmax_{\mathcal{B}_j, \mathfrak{D}_j} f_\text{LLM}(\mathcal{B}_j, \mathfrak{D}_j, \mathcal{X}_\text{con}, S, n_p, \mathcal{C})
\end{equation}

\myparagraph{Layout Generation:} For each frame $B_i \in \mathcal{B}$, we generate a layout $\Lambda_i = \{\mathcal{R}_i, \mathcal{F}_i, \mathcal{G}_i\}$ using exemplar-driven in-context learning with iterative correction (\cref{algo1} in appendix). Here, $\mathcal{R}_i$ are bounding boxes for characters, $\mathcal{F}_i$ are foreground prompts, and $\mathcal{G}_i$ is a background prompt. The layout is refined by minimizing the error between the original and reconstructed prompt $B'_i = \mathcal{A}(\Lambda_i):
\Lambda^*_i = \argmin_{\Lambda_i} \mathcal{E}_\text{rec}\left(B_i, B'_i\right)$
where $\mathcal{E}_\text{rec} = 1 - [\mu_1 \text{sim}_\text{cos} (B_i, B'_i) + \mu_2 \text{sim}_\text{jac} (B_i, B'_i)  - \mu_3 \text{sim}_\text{edit} (B_i, B'_i)]$. Here, $\text{sim}_\text{cos}, \text{sim}_\text{jac}$ and $\text{sim}_\text{edit}$ measure cosine similarity, Jaccard similarity, and edit distance, respectively, with coefficients $\mu_1 = \mu_2 = 1.0$ and $\mu_3$ = 0.01 determined by grid search. $B'_i = \mathcal{A}(\Lambda_i)$ represents the reconstructed prompt, with $\mathcal{A}$ being the layout-to-text generation algorithm from \cite{yin2017obj2text}. This layout optimization improves spatial alignment without model fine-tuning.

\begin{figure*}[!htbp]
    \centering
    \includegraphics[width=\linewidth]{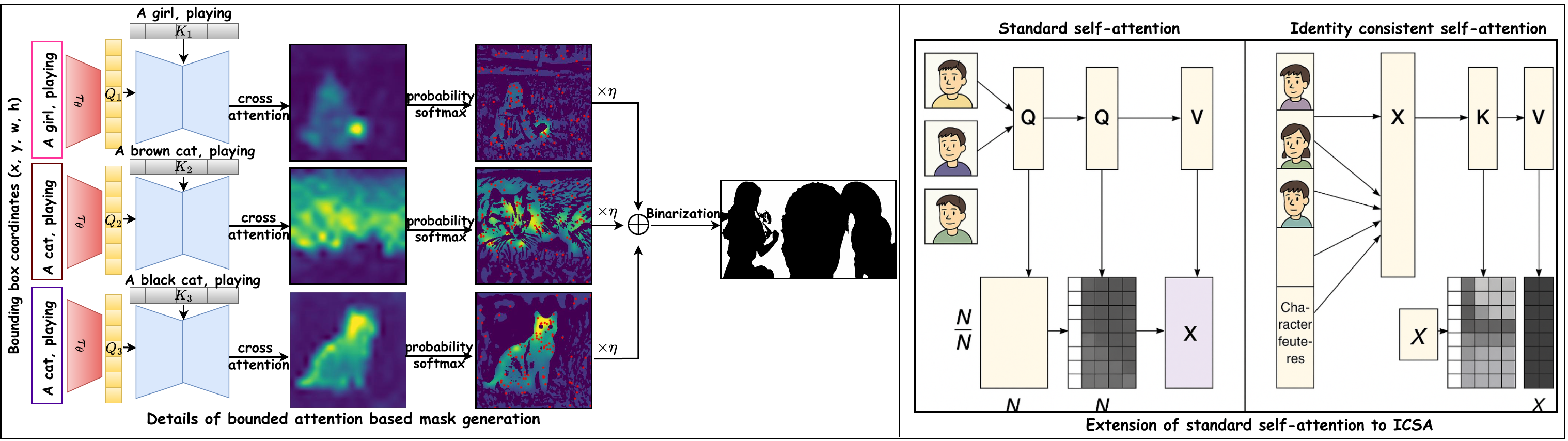}
    \caption{\textbf{Bounded attention based per-box mask (left) and ICSA (right):} Former takes the bounding box of a single character at a time and create binarize mask of the cross attention of $\text{CLIP}_\text{text}$ where the later extends the self-attention by storing the character features in a long vector and update the key and values during frame generation instead of using the same queries like traditional self-attention \cite{zhou2024storydiffusion}. (see \cref{fig:mask1,fig:mask2,fig:mask3,fig:icsa,fig:cross} in the appendix for more details.)}
    \label{fig:mask}
\vspace{-4mm}
\end{figure*}

\subsection{Mask Generation via Bounded Attention}
\label{sec:m2}
Subject-specific masks add flexibility with anisotropic scaling, precise poses, and accurate, versatile shapes and textures over the layout control.
Existing works \cite{wang2023autostory} generate masks from object bounding boxes through dense conditional generation and segmentation with G-SAM \cite{ren2024groundedsam}, but this often introduces artifacts \cite{zhang2023perceptualartifacts}. We address this issue with a bounded attention-based mask generation, preserving subject semantics without fine-tuning. For $n_t$ distinct subjects in the $i$-th frame, we use foreground prompt embeddings $\mathcal{F}_i$ as keys and values, and bounding boxes $\mathcal{R}_i$ as queries, applying an augmented weighting scheme in the diffusion U-Net’s attention layers (see \cref{fig:mask}). Queries $Q_{i,j}$ are derived from region features $R_{i,j}$ such that $Q_{i,j}= \text{GLIGEN}(R_{i,j})$, via the conditional encoder of the GLIGEN \cite{li2023gligen} which encodes spatially aligned conditions (\eg boxes) into the grounding tokens as a query. Similarly, $K_{i,j}$, $V_{i,j}$ come from $\mathcal{F}_{i,j}$ using the CLIP text encoder, $K_{i,j}, V_{i,j}=\text{CLIP}_\text{text}(\mathcal{F}_{i,j})$. The bounded attention map $A_{i,j}$ for the $j$-th character in frame $i$ defined as:
\vspace{-2mm}
\begin{equation}
\vspace{-2mm}
    \label{eq:2}
    A_{(i,j)}=\softmax\left(\dfrac{{Q_{i,j}}^T K_{i,j}}{\sqrt{d}}\right)V_{i,j}
\end{equation}

where $d$ is the affinity between $Q_{i,j}$ and $K_{i,j}$, considering their latent projection dimension. This bounded attention map $A_{(i,j)}$ has been averaged across different heads, layers, and timesteps to obtain $\bar{A}_{(i,j)}$. Then, each pixel of $\bar{A}_{(i,j)}$ has been either classified to background or character to form two different cluster of pixels $P_\text{bg}$ and $P_\text{fg}$ corresponding to the foreground and background pixels, respectively, which we will use in the later stage for iterative mask refinement. Now from the $\bar{A}_{(i,j)}$ we can initialize the bounded mask for each character $j$ in the following way:
\begin{equation}
    M_{(i,j)} = \Norm(\sigmoid(\xi.\Norm(\bar{A}_{(i,j)} - \phi_j))
\end{equation}

Here, $\Norm$ represents the L1-normalization and $\phi_j$ are the hyperparameters for sharpness and soft binarization threshold, respectively. As $M_{(i,j)}$ is obtained with a threshold, it suffers from information flow between characters, generating artifacts. We refined this mask to prevent the information flow with the help of foreground clusters $P_\text{fg}$.
\vspace{-2mm}
\begin{equation}
\vspace{-2mm}
        \tilde{M}_{(i,j)} = \dfrac{M_{(i,j)} \odot P_\text{fg}}{\min(\sum_xM_{(i,j)}[x],\sum_xP_\text{fg}[x])}
\end{equation}
This $\tilde{M}_{(i,j)}$ is obtained subject subject-specific mask which provides precise control over layout. Those $\tilde{M}_{(i,j)}$ are iteratively concatenated for each frame $B_i$:
\vspace{-2mm}
\begin{equation}
\vspace{-2mm}
    \label{eq:3}
    \tilde{M}_{\mathcal{R}_i} = \eta_1 \tilde{M}_{(i,1)} \cup  \eta_2 \tilde{M}_{(i,2)} \cup \ldots \cup \eta_{n_t} \tilde{M}_{(i,n_t)}
\end{equation}
where $\eta_j = \text{area}(\tilde{M}_{(i,j)})/\text{area}(\tilde{M}_{(i,1)} \cup \ldots \cup \tilde{M}_{(i,n_t)})$, ensuring each mask $\tilde{M}_{(i,j)}$ is proportionally weighted in the latent guidance step to form the final mask.

\subsection{Consistent Character Generation}
\label{sec:m3}
We create a character image database $\mathcal{C}_\text{db}$ from character description $C_i \in \mathcal{C}$ using a pretrained diffusion model, serving as a reference for generating consistent latent guidance $Z_\text{fg}$. Consistency is maintained through the proposed ICSA module, and the characters are customized in the masked shape via low-rank adaptation (LoRA) with gradient fusion. This $Z_\text{fg}$ is further denoised with a background latent noise $Z_\text{bg}$ by the diffusion U-Net with RACA to produce final images with consistent characters for each frame $B_i\in \mathcal{B}$. (\textbf{Note:} This $\mathcal{C}_\text{db}$ has been used for LoRA customization we didn't use any additional database for pose customization.)

\begin{figure}[!htbp]
\centering
\includegraphics[width=\linewidth]{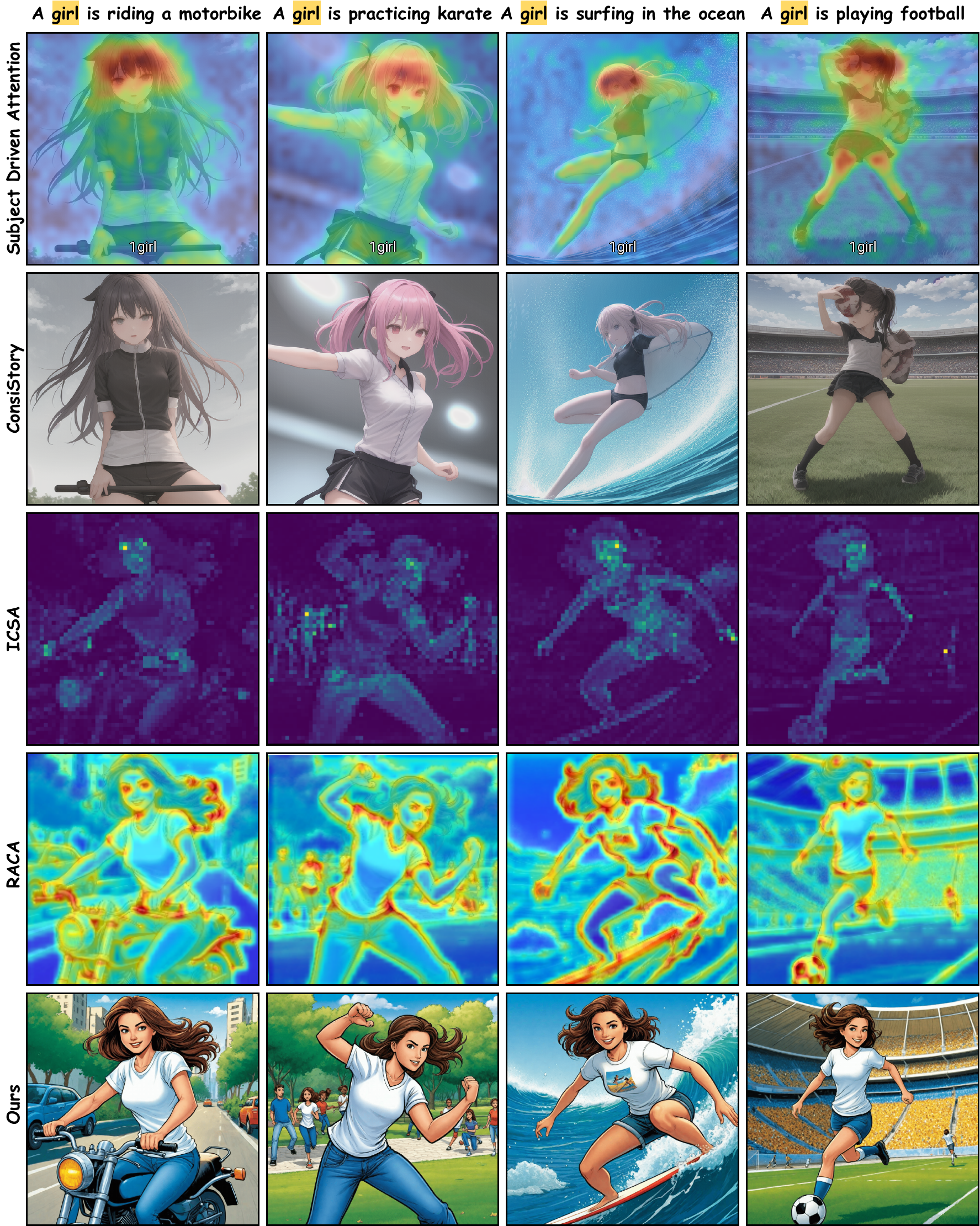}
\caption{\textbf{SDSA \cite{tewel2024training} vs ICSA-RACA:} Former mostly focused on eyes, which leads to artifact generation, and sometimes main characters are missing. In contrast, later puts their primary focus is on the eyes, but also helps to maintain the pose and other attributes.}
\label{fig:ricsa} 
\end{figure}

\myparagraph{Identity-consistent Self-attention (ICSA):}
It extends the self-attention of IP-Adapter \cite{ye2023ip} from $\mathbb{R}^{N \times N}$ to $\mathbb{R}^{N \times n_p N}$, where $n_p$ is the frame count, allowing queries from one foreground latent frames to attend to keys and values across the other foreground latent frames in the batch.
This facilitates shared visual features for repeated objects, improving consistency and supporting character customization in $Z_\text{fg}$ (\cref{fig:ricsa}). Usually, self-attention layers process tokens from a single image patch with linear projections for keys ($W_k$), values ($W_v$), and queries ($W_q$). In ICSA, instead of features from one patch, we concatenate character features $f_i \in \mathcal{C}_\text{db}$ 
to extract tokens $X$. ICSA computes a weighted softmax over $X$ for each character $f_i$ in the $l$-th self-attention layer:
\vspace{-2mm}
\begin{equation}
\vspace{-2mm}
    \label{eq:7}
    \resizebox{0.40\textwidth}{!}{$
    \text{ICSA}^l_{f_i} = \softmax\left(\dfrac{W_q X \cdot {W_k f_i}^T}{\sqrt{d_e}}\right) W_v f_i$}
\end{equation}

where $d_e$ is the feature embedding dimension. This $\text{ICSA}^l_{f_i}$ preserves character consistency, reduces subject interference, and customizes them collectively in $Z_\text{fg}$ as in \cref{eq:6}. (see \cref{sec:RICSA} in the appendix for more details.)

\begin{figure}[!htbp]
\centering
\includegraphics[width=\linewidth]{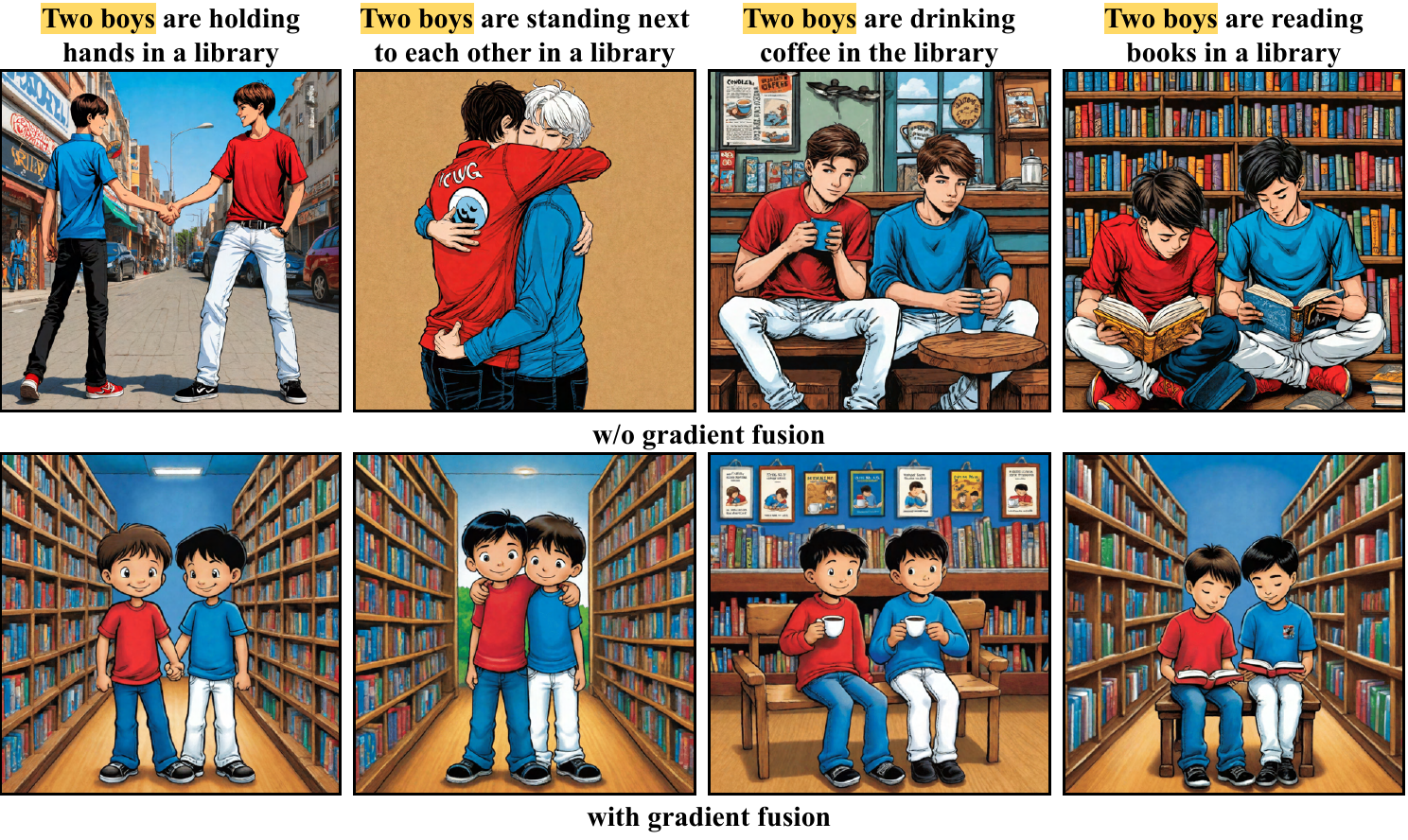}
\caption{\textbf{Gradient Fusion (GF):} The character's attributes can be inconsistent (white hair in the second frame) while GF resolves those inconsistencies in the final frame generation.}
\label{fig:gf}
\end{figure}

\myparagraph{Foreground Latent Generation with LoRA:}
To generate $Z_\text{fg}$, character features $f_i\in \mathcal{C}_\text{db}$ alone are insufficient, as their poses must align with mask $\tilde{M}_{\mathcal{R}_i}$. This alignment is achieved through LoRA, while ICSA preserves character identity and allows flexible multi-character customization with gradient fusion. Given a reference character $f_i \in \mathcal{C}_\text{db}$, a pretrained IP-adapter \cite{ye2023ip} with ICSA is fine-tuned using LoRA to capture detailed subject characteristics. Each character is generated in different ranks ($\sigma$) to maintain consistency of identity with ICSA and preservation of character details. The rank of the LoRA ($\sigma$) has been set to 256 empirically to mitigate the trade-off between the precision of character details (higher rank $\rightarrow$ better precision), and computational trade-off (low-rank $\rightarrow$ faster computation). These character face attributes are treated as different concepts to learn via LoRA and customizing the guided mask through gradient fusion \cite{gu2024mix} to produce $Z_\text{fg}$ (see \cref{fig:gf}):
\begin{equation}
\label{eq:4}
Z_\text{fg} = E_{\tilde{M}_{\mathcal{R}_i}, \mathcal{C}_\text{db}, t}|| \theta(\tilde{M}_{\mathcal{R}_i}, \mathcal{C}_\text{db}, t) - \theta_{\sigma}(\tilde{M}_{\mathcal{R}_i}, \mathcal{C}_\text{db}, t) ||^2
\end{equation}
$E_{\tilde{M}_{\mathcal{R}_i}, \mathcal{C}_\text{db}, t}$ is the IP-adapter energy function to optimize, and $\theta$ is the classifier-free guidance:
\vspace{-2mm}
\begin{equation}
\vspace{-2mm}
\label{eq:5}
\theta(\tilde{M}_{\mathcal{R}_i}, \mathcal{C}_\text{db}, t) = \sum_{l=1}^L \bigoplus_{f_i \in \mathcal{C}_\text{db}} (\tilde{M}_{\mathcal{R}_i} \odot Z_{o_t} \odot \text{ICSA}^l_{f_i})
\end{equation}
Here, $\bigoplus$ denotes concatenation and $L$ is the no. of self-attention layers, and $Z_{o_t}$ is the Gaussian noise in the latent space at timestep $t$. This classifier-free guidance $\theta$ helps to adapt the U-Net from conditional to unconditional weight update with LoRA through gradient fusion:
\vspace{-4mm}
\begin{multline} \label{eq:6}
\vspace{-4mm}
\theta_\sigma (\tilde{M}_{\mathcal{R}_i}, \mathcal{C}_\text{db}, t) = \Delta \omega \sum_{l=1}^L \bigoplus_{f_i \in \mathcal{C}_\text{db}}  \left( \tilde{M}_{\mathcal{R}_i} \odot Z_{o_t} \odot \text{ICSA}^l_{f_i}\right) \\ + (1 - \Delta \omega )\frac{1}{L}\sum_{l=1}^L\bigoplus_{f_i \in \mathcal{C}_\text{db}} 2\sigma \left(Z_{o_t} \odot \text{ICSA}^l_{f_i}\right)
\end{multline}
where $\Delta \omega$ denotes updated weights in the diffusion U-Net during latent guidance.

\myparagraph{Region-aware Cross-attention (RACA):} Once the latent guidance $Z_\text{fg}$ is computed, 
we pass it through another diffusion U-Net to generate a coherent background aligned with $\mathcal{C}$. During the denoising of this background latent, character consistency is maintained using ICSA. However, cross-attention leakage can disrupt object positioning, resulting in image distortion.
To address this, we employ region-aware cross-attention (RACA), a normalized sigmoid variant of cross-attention \cite{tang2022daam}, as shown in \cref{fig:cross} in the appendix.

\myparagraph{Background Latent Denoising:}
After generating the foreground latent $Z_\text{fg}$, we further encode it via an image adapter, and further denoise it with The background latent $Z_{\text{bg}_t}$, preserving the character attributes, to generate a coherent background to make the scene realistic by diminishing the transition effect between foreground and background:
\begin{equation}
\label{eq:new}
\resizebox{0.85\linewidth}{!}{$
    Z_{\text{bg}_{t-1}} = Z_{\text{bg}_t} - \lambda \nabla_{Z_{\text{bg}_t}} \sum_{h_i \in H}  (Z_{\text{bg}_t} \odot h_i) + (Z_\text{fg} \odot \text{RACA}^l_i)$}
\end{equation}

$H$ is the $\text{CLIP}_\text{text}$ encoded background caption, and $\lambda$ is the subject guidance factor.

\subsection{Dialogue Rendering}
\label{sec:m5}
The final step of TaleDiffusion is to render dialogues $\mathcal{D}_i$ into each frame $B_i$, placing them in bubbles and assigning them to corresponding characters $C_i \in \mathcal{C}$ via the dialogue renderer $R_T$. This is the first method, to our knowledge, that assigns dialogue bubbles directly to characters, enhancing comic realism. Using CLIPSeg \cite{luddecke2022image}, we locate each bubble relative to the character’s head based on the specifications in $C_i$. If the character’s head is on the left side of the frame, the bubble is positioned on the right, and vice versa, ensuring it stays within the image boundaries. To ensure bubbles fit within the frame, dialogues are split into lines of up to a certain number of words, determining bubble width and height, with font size and style adjusted to fit. We position the bubble precisely by extracting the head’s edge location and using CLIPSeg logits to locate the arrow coordinates $(x,y)$. We iteratively check for any bubble's overlap with the character’s face (\cref{algo2,algo3} in the appendix), adjusting the bubble position as needed. This approach keeps bubbles close to characters without obscuring any details, enhancing interaction within each frame without conflicts.

\section{Experimental Results and Analysis}
\label{sec:exp}
\begin{figure}[!htbp]
\vspace{-2mm}
\centering
\includegraphics[width=\linewidth]{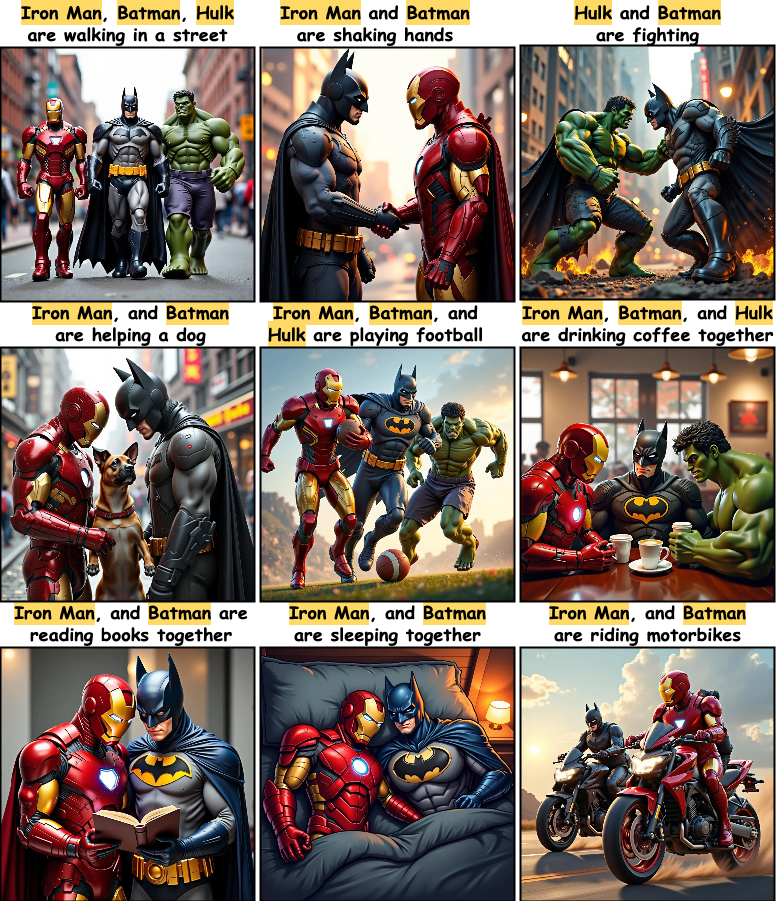}
\caption{TaleDiffusion improves interactivity by dynamically adjusting character relationships and environmental interactions.}
\vspace{-4mm}
\label{fig:int}
\end{figure}

\myparagraph{Experiments on Interactivity:} As depicted in \cref{fig:int}, TaleDiffusion makes panels feel alive by updating how characters relate to each other and their surroundings as the story progresses, rather than treating them as isolated cutouts \cite{he2024dreamstory,tao2024storyimager,zhu2023cogcartoon}. Concretely, it localizes each character with bounded, per-box attention masks so interactions are composed deliberately instead of bleeding together, then uses identity-consistent self-attention (ICSA) and region-aware cross-attention (RACA) to keep identities stable while adjusting poses and spatial relations to match the evolving scene (i.e., true character-to-character and character-to-environment interplay). This is paired with CLIPSeg-based dialogue rendering that anchors each speech bubble to the correct speaker’s head, so conversational turns are visually grounded and responsive within the frame. Together, these components yield interactive behavior missing in prior systems, which often drop objects, misplace characters, or misassign dialogue because they lack mask-guided control and character-aware attention.

\subsection{Qualitative Evaluation}
\label{sec:qual}

\begin{figure*}[!htbp]
\centering
\includegraphics[width=\textwidth]{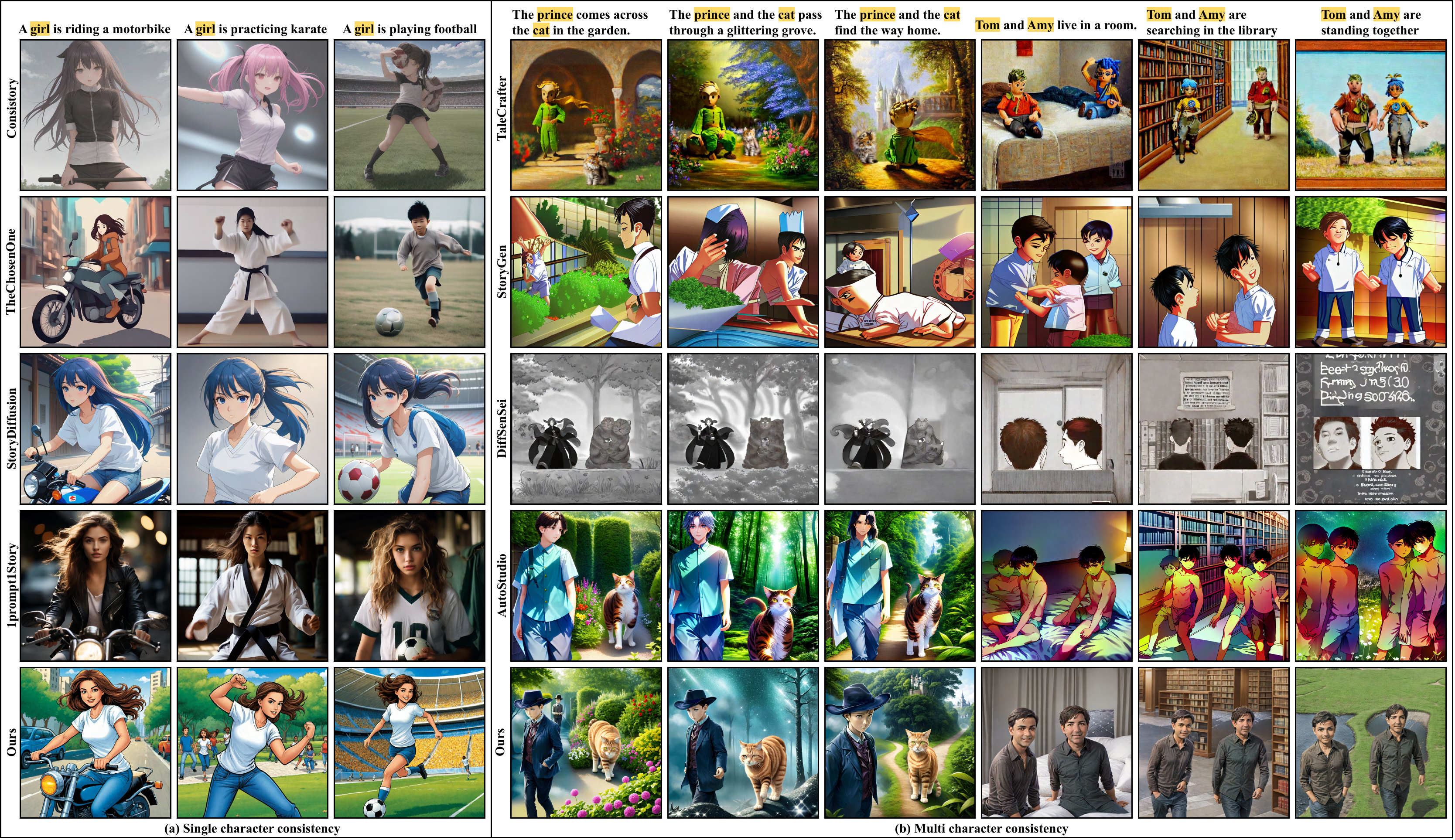}
\caption{TaleDiffusion outperforms the prior works by maintaining the consistency and spatial relationship between multiple objects.}
\label{fig:comp}
\vspace{-4mm}
\end{figure*}

We compare TaleDiffusion with single-character methods \cite{avrahami2024chosen,tewel2024training,liuone}, two-character methods \cite{talecrafter2024,liu2024intelligent,zhou2024storydiffusion} as well as multi character methods \cite{cheng2024autostudio,wu2024diffsensei} in \cref{fig:comp}(a) and (b), respectively. For multi-character story visualization ($\geq 2$), please refer to \cref{sec:lecun} in the suppl. As competitor methods typically fail to generate more than two characters per panel, including the multi-character methods \cite{cheng2024autostudio,wu2024diffsensei}. \cref{fig:comp} only includes stories with a maximum of two characters per panel. The results in \cref{fig:comp} show that TaleDiffusion consistently produces panels aligned with text prompts and with fewer artifacts, excelling in the story generation.

As shown in \cref{fig:comp}(a), ConsiStory \cite{tewel2024training} can consistently generate characters but often omits auxiliary objects (\eg, motorbike). TheChosenOne \cite{avrahami2024chosen} improves text-to-image similarity over ConsiStory, but struggles with character consistency. StoryGen \cite{liu2024intelligent} often fails to follow prompts, lacks consistency, and generates artifacts for both single and multiple characters. StoryDiffusion \cite{zhou2024storydiffusion} and 1prompt1story \cite{liuone} improve character consistency for single characters; they still generate artifacts and fail to maintain interactions with objects as described in the text prompts. Moreover, they struggle to consistently generate two human characters in a single frame (see \cref{fig:comp} (b)). TaleCrafter \cite{talecrafter2024} adheres to text prompts; it suffers from poor image quality and significant noise. DiffSensei \cite{wu2024diffsensei} neither has character consistency nor follows the text prompt. Although AutoStudio \cite{cheng2024autostudio} improves the character consistency generates the largest number of artifacts.

In contrast, TaleDiffusion generates consistent, artifact-free characters for stories with both single and multiple characters by utilizing the proposed mask guidance to prevent artifacts during the denoising process. ICSA ensures character consistency in latent guidance $Z_\text{fg}$ and background $Z_\text{bg}$ through IP-Adapter \cite{ye2023ip}, while also controlling position and size to maintain consistency. Additionally, TaleDiffusion effectively preserves interactions between multiple characters and background objects, as shown in \cref{fig:comp} (more comparisons are in \cref{sec:baseline} of the appendix).

\subsection{Quantitative Evaluation}
\label{sec:quan}
We quantitatively assess the performance of TaleDiffusion in \cref{tab:1} using various metrics: Artifacts Score (with a classifier trained to detect artifacts; see \cref{sec:artifacts} in the suppl.), average character consistency (aCCS) \cite{zhou2024storydiffusion} based on cosine similarity across characters in different panels, VQAScore \cite{lin2025evaluating} for text-to-image alignment, average aesthetic score (aAes) \cite{Aes2022}, and average Human Performance Score (aHPS) \cite{wu2023human} for perceptual quality assessment, comparing TaleDiffusion with existing methods.

\begin{table}[!htbp]
\centering
\caption{Evaluation with existing methods}
\label{tab:1}
\resizebox{\linewidth}{!}{
\begin{tabular}{clccccc}
\hline
 &
  Method/Metrics &
  Artifacts Score $\downarrow$ &
  aCCS \cite{zhou2024storydiffusion} $\uparrow$ &
  VQAScore \cite{lin2025evaluating} $\uparrow$ &
  aAes. \cite{Aes2022}$\uparrow$ &
  aHPS \cite{wu2023human}$\uparrow$ \\ \hline
\multirow{7}{*}{\rotatebox[origin=c]{90}{\textbf{Single char.}}} &
  TheChosenOne \cite{avrahami2024chosen} &
  0.6237 &
  0.6432 &
  0.5234 &
  3.2144 &
  0.1997 \\
 &
  Consistory \cite{tewel2024training} &
  0.5992 &
  0.6737 &
  0.5123 &
  5.5312 &
  0.2617 \\
 &
  StoryGen \cite{liu2024intelligent} &
  0.8011 &
  0.5698 &
  0.6312 &
  2.9182 &
  0.1892 \\
 &
  StoryDiffusion \cite{zhou2024storydiffusion} &
  0.5432 &
  0.7487 &
  0.7321 &
  6.5432 &
  0.2888 \\
 &
  StoryAdapter \cite{Mao2024StoryAdapterAT} &
  0.6141 &
  0.7243 &
  0. 5938 &
  5.4712 &
  0.2347 \\
 &
  1prompt1story \cite{liuone} &
  0.3236 &
  0.7614 &
  0.7819 &
  6.7717 &
  0.3003 \\
 &
  TaleDiffusion (Ours) &
  \textbf{0.1021} &
  \textbf{0.8088} &
  \textbf{0.8322} &
  \textbf{6.7832} &
  \textbf{0.3276} \\ \hline
\multirow{8}{*}{\rotatebox[origin=c]{90}{\textbf{Multi char.}}} &
  TaleCrafter \cite{talecrafter2024} &
  0.3245 &
  0.6758 &
  0.6976 &
  4.1242 &
  0.2131 \\
 &
  StoryGen \cite{liu2024intelligent} &
  0.8873 &
  0.4392 &
  0.3091 &
  3.7121 &
  0.1791 \\
 &
  StoryDiffusion \cite{zhou2024storydiffusion} &
  0.2737 &
  0.6891 &
  0.4076 &
  5.7135 &
  0.2772 \\
 &
  StoryAdapter \cite{Mao2024StoryAdapterAT} &
  0.4194 &
  0.6482 &
  0.5013 &
  4.8821 &
  0.2195 \\
 &
  1prompt1story \cite{liuone} &
  0.2731 &
  0.7657 &
  0.4415 &
  6.1121 &
  0.2838 \\
 &
  AutoStudio \cite{cheng2024autostudio} &
  0.2387 &
  0.7543 &
  0.6427 &
  6.2143 &
  0.2912 \\
 &
  DiffSenSei \cite{wu2024diffsensei} &
  0.4728 &
  0.4112 &
  0.3941 &
  3.6139 &
  0.2017 \\
 &
  TaleDiffusion (Ours) &
  \textbf{0.1103} &
  \textbf{0.7992} &
  \textbf{0.7864} &
  \textbf{6.5412} &
  \textbf{0.3301} \\ \hline
\end{tabular}
}
\end{table}

In \cref{tab:1}, we present results for two settings: single-character and multi-character, each using 60 unique prompts across 10 stories (i.e., 6 prompts per story) with 10 panels per prompt. StoryGen \cite{liu2024intelligent} produced the most artifacts during denoising in both single-character ($\sim$80\%) and multi-character ($\sim$88\%) settings. Additionally, low aCCS and VQAScore suggest that StoryGen struggles with consistency and prompt alignment. StoryDiffusion \cite{zhou2024storydiffusion} also showed a notable drop in aCCS and VQAScore when moving from single to multiple characters, indicating poor consistency. While TaleDiffusion experienced minor performance declines from single to multiple characters due to increased internal noise, it achieved high aCCS and aHPS scores, indicating better diversity and perceptual quality, and outperformed all existing methods across all metrics by a significant margin.

\begin{table}
\centering
\caption{Part-by-part analysis of TaleDiffusion}
\label{tab:2}
\resizebox{\linewidth}{!}{
\begin{tabular}{ccccccc}
\hline
Method/Metrics &
  Artifacts Score $\downarrow$ &
  aCCS \cite{zhou2024storydiffusion} $\uparrow$ &
  VQAScore \cite{lin2025evaluating} $\uparrow$ &
  aAes. \cite{Aes2022}$\uparrow$ &
  aHPS \cite{wu2023human}$\uparrow$ &
  R@(\#text) \cite{vivoli2024comix} $\uparrow$ \\ \hline
Baseline (IP-Adapter \cite{ye2023ip}) & 0.8032          & 0.3012          & 0.2813          & 2.8713          & 0.1029          & -               \\
+ layout                                               & 0.7013          & 0.3214          & 0.3011          & 3.1476          & 0.1021          & -               \\
+ Mask                                                 & 0.4783          & 0.3471          & 0.3617          & 3.2214          & 0.1192          & -               \\
+ Bounded attention mask                               & 0.2013          & 0.3567          & 0.3813          & 3.6392          & 0.1310          & -               \\
+ ICSA1                                                & 0.1672          & 0.6581          & 0.6712          & 4.1278          & 0.2117          & -               \\
+ ICSA2                                                & 0.1212          & 0.7212          & 0.7331          & 5.7142          & 0.2737          & -               \\
+ RACA                                                 & \textbf{0.1103} & \textbf{0.7992} & \textbf{0.7864} & \textbf{6.5412} & \textbf{0.3301} & -               \\ \hline
TaleDiffusion (fg + bg)                                & 0.3016          & 0.6931          & 0.6212          & 5.2123          & 0.2817          & -               \\
TaleDiffusion (fg $\rightarrow$ bg)                    & \textbf{0.1103} & \textbf{0.7992} & \textbf{0.7864} & \textbf{6.5412} & \textbf{0.3301} & -               \\ \hline
w/o CLIPSeg                                            & 0.1103          & 0.7992          & 0.7864          & 6.5412          & 0.3301          & 0.5721          \\
w/ CLIPSeg                                             & 0.1103          & 0.7992          & 0.7864          & 6.5412          & 0.3301          & \textbf{0.6413} \\
DiffSenSei \cite{wu2024diffsensei}    & 0.4728          & 0.4112          & 0.3941          & 3.6139          & 0.2017          & 0.2054          \\ \hline
\end{tabular}
}
\end{table}

\subsection{Ablation Study}
\label{sec:ab}

\myparagraph{Story Expansion Module:}
This experiment aims to identify the optimal LLM for generating complete panel-wise descriptions and coherent dialogues based on user-provided themes. We utilized the eBDtheque \cite{guerin2013ebdtheque}, a French comic dataset with 100 pages, summarizing each page using GPT-4. We then tried to generate complete panel descriptions with dialogues using various LLMs, as reported in \cref{tab:04}. Results indicate that GPT4 \cite{achiam2023gpt} outperforms other models (Llama 7B \cite{touvron2023llama}, Vicuna-13B \cite{chiang2023vicuna}, GPT-3.5 \cite{huggingfaceXenovagpt35turboHugging}, Claude-2 \cite{huggingfaceTheBlokeclaude2alpaca13BGPTQHugging}, Claude-3.5 Sonnet \cite{anthropicClaudeSonnet}) in terms of story coherence and creativity. While Claude-3.5 Sonnet showed better readability and completeness, GPT-4 led overall. Consequently, we chose GPT-4 with in-context learning as our story generator for further experiments (see \cref{tab:03}).
(Note: Readability, Completeness, and Coherency score were evaluated using TextDescriptives \cite{githubGitHubHLasseTextDescriptives} Library, and Creativity was assessed using the SAI metric \cite{organisciak2023beyond}.)

\begin{table*}[!htbp]
\centering
\caption{Evaluation of story generation with LLMs.}
\label{tab:03}
\resizebox{\textwidth}{!}{
\begin{tabular}{l|ccccc|ccccc}
\hline
\multirow{2}{*}{Methods/Metrics} & \multicolumn{5}{c|}{chain-of-thoughts \cite{zhang2023multimodal}} & \multicolumn{5}{c}{in-context learning \cite{wang2024large}} \\ \cline{2-11} 
 &
  \multicolumn{1}{c|}{Readability $\uparrow$} &
  \multicolumn{1}{c|}{Coherence $\uparrow$} &
  \multicolumn{1}{c|}{Creativity $\uparrow$} &
  \multicolumn{1}{c|}{Completeness $\uparrow$} &
  Average $\uparrow$&
  \multicolumn{1}{c|}{Readability $\uparrow$} &
  \multicolumn{1}{c|}{Coherence $\uparrow$} &
  \multicolumn{1}{c|}{Creativity $\uparrow$} &
  \multicolumn{1}{c|}{Completeness $\uparrow$} &
  Average $\uparrow$\\ \hline
Llama-7B  \cite{touvron2023llama}                       & 0.6428   & 0.5117   & 0.4392   & 0.5017  & 0.5238  & 0.7824   & 0.6711   & 0.5923   & 0.6170  & 0.6657  \\
Vicuna-13B  \cite{chiang2023vicuna}                     & 0.6178   & 0.5288   & 0.5651   & 0.5255  & 0.5593  & 0.7871   & 0.6882   & 0.6156   & 0.6552  & 0.6865  \\
GPT-3.5 \footnote[1]{}                         & 0.6119   & 0.6514   & 0.5493   & 0.6312  &  0.6872  & 0.6110   & 0.7145   & 0.6934   & 0.7123  &  0.7278 \\
GPT-4 \cite{achiam2023gpt}                           & 0.7121   & \textbf{0.7153}   & \textbf{0.7020}   & \textbf{0.7342}  & \textbf{0.7160}  & 0.8036   & \textbf{0.8013}   & \textbf{0.7811}   & 0.8196  & \textbf{0.8014}  \\
Claude-2 \footnote[2]{}                        & 0.6080   & 0.5219   & 0.5127   & 0.5277  & 0.5425   & 0.7800   & 0.6912   & 0.6721   & 0.6772  & 0.7051  \\
Claude-3.5 Sonnet \footnote[3]{}               & \textbf{0.7282}   & 0.7021   & 0.6924   & 0.7132  & 0.7089  & \textbf{0.8067}   & 0.7727   & 0.7467   & \textbf{0.8211}  & 0.7868  \\ \hline
\end{tabular}
}
\label{tab:04}
\vspace{-4mm}
\end{table*}

\begin{figure}[!htbp]
\centering
\includegraphics[width=\linewidth]{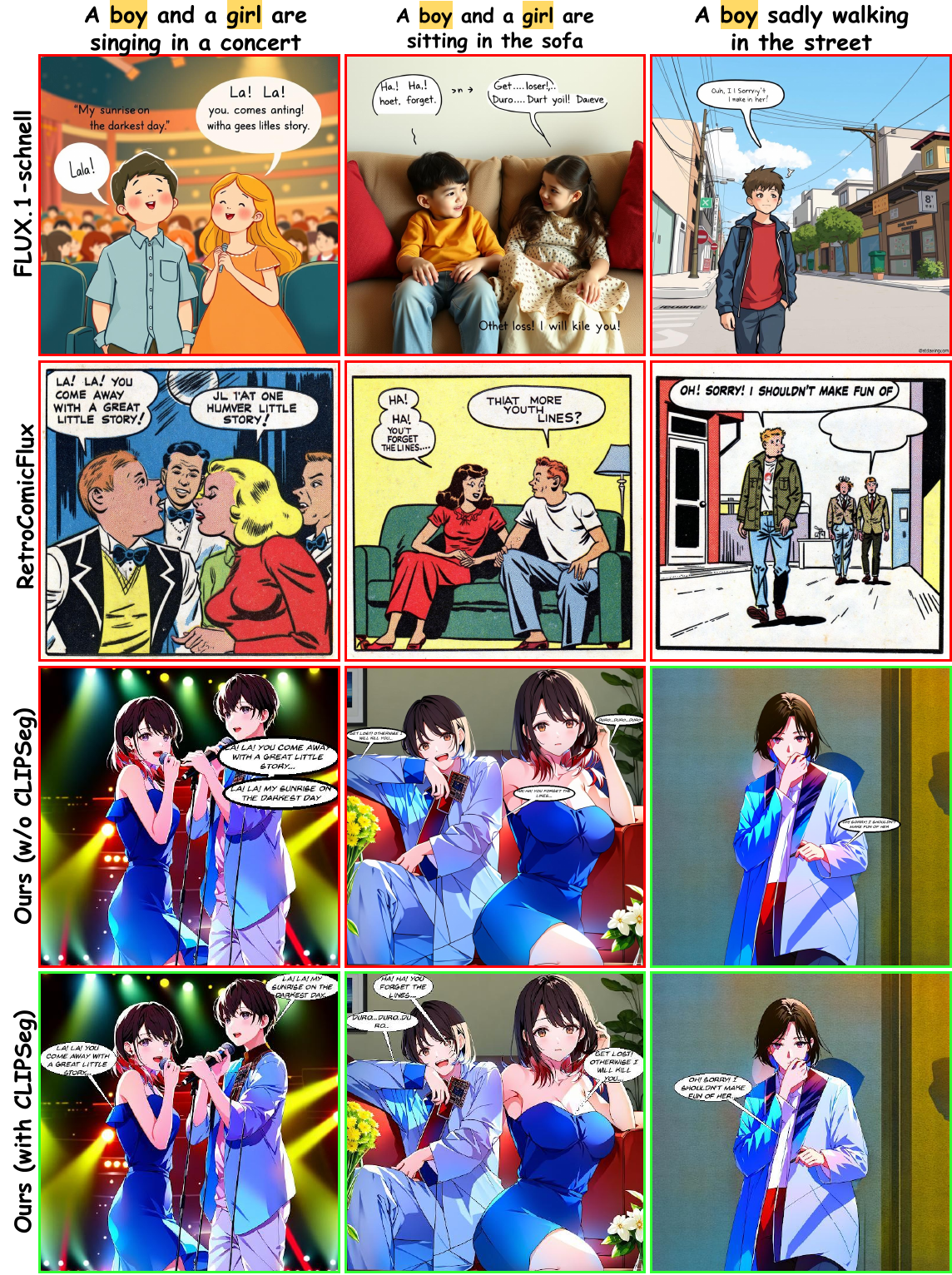}
\caption{\textbf{CLIPSeg in dialogue rendering:} FLUX.1-schnell and RetroComicFlux fail to place text in bubbles and assign multiple bubbles to one character. In contrast, our method precisely places text and uses CLIPSeg to correctly assign bubbles to characters.}
\label{fig:abd}
\end{figure}

\myparagraph{Dialogue Rendering:}
We evaluate CLIPSeg’s \cite{lin2025evaluating} ability to link dialogues to the correct characters using the R@\#text metric \cite{vivoli2024comix}, with LLM-generated dialogues as ground truth. As shown in \cref{tab:2}, CLIPSeg improves bubble assignment accuracy by $\sim$ 7\% via character-aware segmentation. Compared to FLUX.1-schnell \cite{huggingfaceBlackforestlabsFLUX1schnellHugging} and RetroComicFlux \cite{huggingfaceRenderartistretrocomicfluxHugging} (\cref{fig:abd}), CLIPSeg is more reliable—others struggle with text placement or generate redundant bubbles. FLUX.1’s reliance on T5 embeddings limits its precision.

\begin{figure}[!htbp]
\centering
\includegraphics[width=\linewidth]{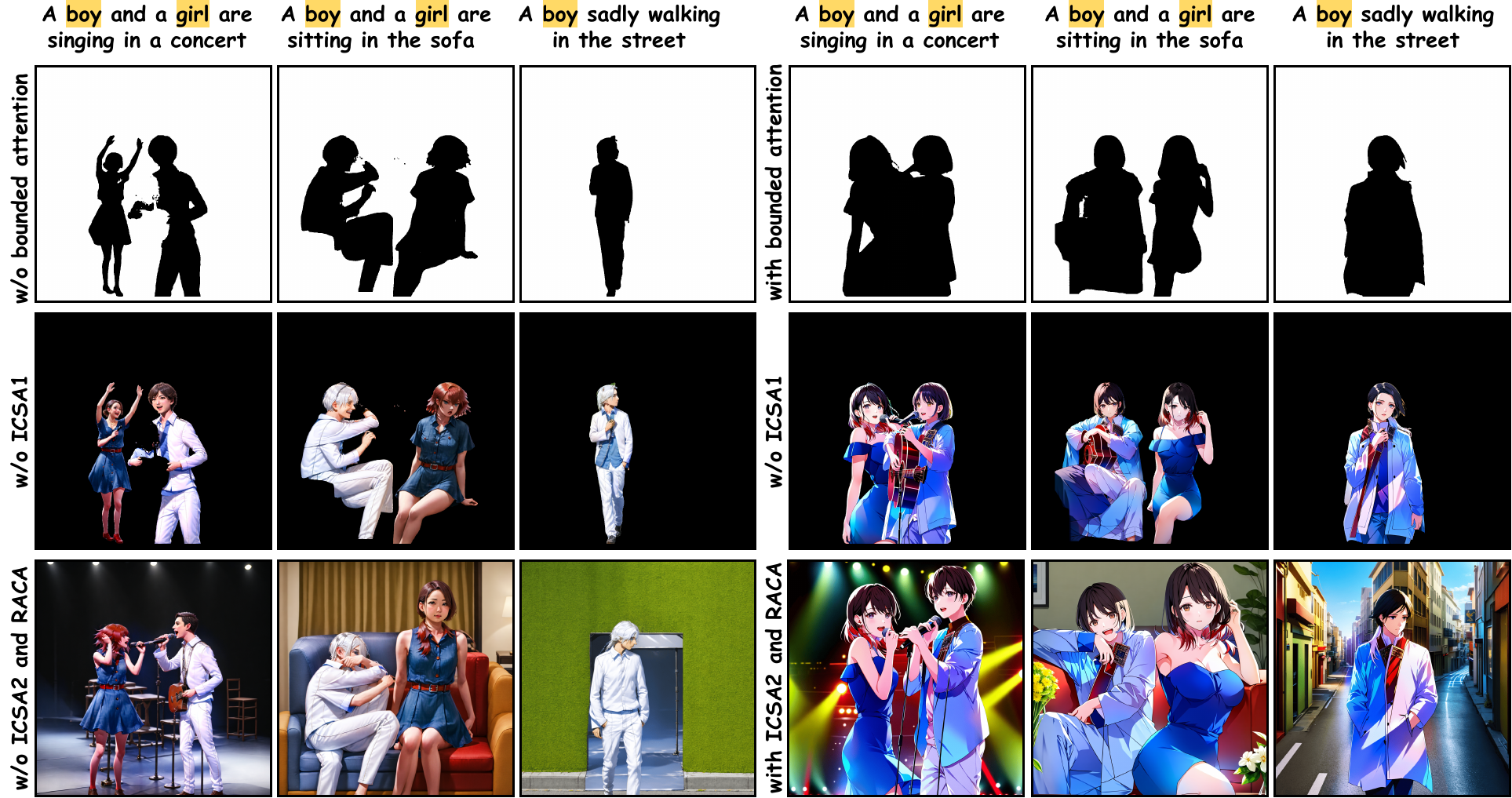}
\caption{\textbf{Ablation Study:} While bounded attention mitigates the artifacts generation, ICSA and RACA improve character consistency by maintaining their spatial relationship.}
\label{fig:new_ab}
\vspace{-4mm}
\end{figure}

\myparagraph{Story Visualization:}
This experiment demonstrates how guidance mechanisms improve multi-character story visualization over the IP-Adapter \cite{ye2023ip}, enhancing consistency, positioning, and reducing artifacts (see \cref{tab:2}). Non-uniform scaling distorts character masks (\cref{fig:new_ab}, row 1, columns 1–3). To address this, we use bounded attention (\cref{eq:2}), cutting artifacts by $\sim$ 60\%. For character consistency, ICSA improves results by $\sim$ 30\%. As characters may shift during background denoising (\cref{fig:new_ab}, row 3, columns 1–3), ICSA is applied twice—during latent guidance and background generation. RACA ensures correct positioning and realism (\cref{fig:new_ab}, row 3, columns 4–6) via normalization.

\section{Conclusion}
\label{sec:conc}
We proposed TaleDiffusion, a framework for generating realistic stories with multiple characters, addressing key limitations in existing story generation methods. By introducing ICSA and RACA, TaleDiffusion ensured character consistency and spatial alignment across panels, maintaining complex multi-character narratives. Mask-guided latent generation and bounded attention significantly reduced artifacts, resulting in cleaner and more controlled outputs. Additionally, TaleDiffusion provided accurate dialogue rendering, uniquely associating dialogue bubbles with individual characters using CLIPSeg. Although multi-character customization with LoRA impacted inference time (see \cref{sec:fail} in suppl.), experimental results demonstrated TaleDiffusion’s superior performance in generating artifact-free, consistent comics aligned with narrative prompts, marking an advancement in story generation.

{\small
\bibliographystyle{ieee_fullname}
\bibliography{main}

@String(CVPR= {IEEE Conf. Comput. Vis. Pattern Recog.})

@String(ICCV= {Int. Conf. Comput. Vis.})

@String(ECCV= {Eur. Conf. Comput. Vis.})

@String(TOG= {ACM Trans. Graph.})

@String(CVPR  = {CVPR})

@String(ICCV  = {ICCV})

@String(ECCV  = {ECCV})

@String(TOG   = {ACM TOG})

@article{tatalovi2009scienceca,
  title={Science comics as tools for science education and communication: a brief, exploratory study},
  author={Mi{\'c}o Tatalovi{\'c}},
  journal={JCOM},
  volume={8},
  number={4},
  year={2009}
}

@article{hosler2011comics,
    author = {Hosler, Jay and Boomer, K. B.},
    title = {Are Comic Books an Effective Way to Engage Nonmajors in Learning and Appreciating Science?},
    journal = {CBE—Life Sciences Education},
    year = {2011}
}

@inproceedings{goodfellow2014gans,
    author = {Ian J. Goodfellow and Jean Pouget-Abadie and Mehdi Mirza and Bing Xu and David Warde-Farley and Sherjil Ozair and Aaron Courville and Yoshua Bengio},
    title = {Generative Adversarial Networks}, 
    booktitle = {NeurIPS},
    year = {2014}
}

@inproceedings{rombach2022sd,
    author = {Robin Rombach and Andreas Blattmann and Dominik Lorenz and Patrick Esser and Björn Ommer},
    title = {High-Resolution Image Synthesis with Latent Diffusion Models}, 
    booktitle = {CVPR},
    year = {2022}
}

@inproceedings{liu2024intelligent,
  title={Intelligent Grimm-Open-ended Visual Storytelling via Latent Diffusion Models},
  author={Liu, Chang and Wu, Haoning and Zhong, Yujie and Zhang, Xiaoyun and Wang, Yanfeng and Xie, Weidi},
  booktitle={Proceedings of the IEEE/CVF Conference on Computer Vision and Pattern Recognition},
  pages={6190--6200},
  year={2024}
}

@article{ren2024groundedsam,
  title={Grounded SAM: Assembling Open-World Models for Diverse Visual Tasks}, 
  author={Tianhe Ren and Shilong Liu and Ailing Zeng and Jing Lin and Kunchang Li and He Cao and Jiayu Chen and Xinyu Huang and Yukang Chen and Feng Yan and Zhaoyang Zeng and Hao Zhang and Feng Li and Jie Yang and Hongyang Li and Qing Jiang and Lei Zhang},
  year={2024},
  journal={arXiv}
}

@inproceedings{zhang2023perceptualartifacts,
  title={Perceptual Artifacts Localization for Image Synthesis Tasks}, 
  author={Lingzhi Zhang and Zhengjie Xu and Connelly Barnes and Yuqian Zhou and Qing Liu and He Zhang and Sohrab Amirghodsi and Zhe Lin and Eli Shechtman and Jianbo Shi},
  booktitle={ICCV},
  year={2023}  
}

@article{tewel2024training,
  title={Training-free consistent text-to-image generation},
  author={Tewel, Yoad and Kaduri, Omri and Gal, Rinon and Kasten, Yoni and Wolf, Lior and Chechik, Gal and Atzmon, Yuval},
  journal={ACM Transactions on Graphics (TOG)},
  volume={43},
  number={4},
  pages={1--18},
  year={2024},
  publisher={ACM New York, NY, USA}
}

@inproceedings{avrahami2024chosen,
  title={The chosen one: Consistent characters in text-to-image diffusion models},
  author={Avrahami, Omri and Hertz, Amir and Vinker, Yael and Arar, Moab and Fruchter, Shlomi and Fried, Ohad and Cohen-Or, Daniel and Lischinski, Dani},
  booktitle={ACM SIGGRAPH 2024 Conference Papers},
  pages={1--12},
  year={2024}
}

@article{cheng2024autostudio,
  title={AutoStudio: Crafting Consistent Subjects in Multi-turn Interactive Image Generation},
  author={Cheng, Junhao and Lu, Xi and Li, Hanhui and Zai, Khun Loun and Yin, Baiqiao and Cheng, Yuhao and Yan, Yiqiang and Liang, Xiaodan},
  journal={arXiv preprint arXiv:2406.01388},
  year={2024}
}

@article{cheng2024theatergen,
  title={TheaterGen: Character Management with LLM for Consistent Multi-turn Image Generation},
  author={Cheng, Junhao and Yin, Baiqiao and Cai, Kaixin and Huang, Minbin and Li, Hanhui and He, Yuxin and Lu, Xi and Li, Yue and Li, Yifei and Cheng, Yuhao and others},
  journal={arXiv preprint arXiv:2404.18919},
  year={2024}
}

@article{zhou2024storydiffusion,
  title={StoryDiffusion: Consistent Self-Attention for Long-Range Image and Video Generation},
  author={Zhou, Yupeng and Zhou, Daquan and Cheng, Ming-Ming and Feng, Jiashi and Hou, Qibin},
  journal={arXiv preprint arXiv:2405.01434},
  year={2024}
}

@article{shen2023storygpt,
  title={StoryGPT-V: Large Language Models as Consistent Story Visualizers},
  author={Shen, Xiaoqian and Elhoseiny, Mohamed},
  year={2023},
  journal={arXiv}
}

@inproceedings{Mao2024StoryAdapterAT,
  title={Story-Adapter: A Training-free Iterative Framework for Long Story Visualization},
  author={Jiawei Mao and Xiaoke Huang and Yunfei Xie and Yuanqi Chang and Mude Hui and Bingjie Xu and Yuyin Zhou},
  booktitle={arXiv},
  year={2024}  
}

@article{tao2024storyimager,
  title={StoryImager: A Unified and Efficient Framework for Coherent Story Visualization and Completion},
  author={Tao, Ming and Bao, Bing-Kun and Tang, Hao and Wang, Yaowei and Xu, Changsheng},
  journal={arXiv preprint arXiv:2404.05979},
  year={2024}
}

@inproceedings{rahman2023make,
  title={Make-a-story: Visual memory conditioned consistent story generation},
  author={Rahman, Tanzila and Lee, Hsin-Ying and Ren, Jian and Tulyakov, Sergey and Mahajan, Shweta and Sigal, Leonid},
  booktitle={Proceedings of the IEEE/CVF Conference on Computer Vision and Pattern Recognition},
  pages={2493--2502},
  year={2023}
}

@inproceedings{pan2024synthesizing,
  title={Synthesizing coherent story with auto-regressive latent diffusion models},
  author={Pan, Xichen and Qin, Pengda and Li, Yuhong and Xue, Hui and Chen, Wenhu},
  booktitle={Proceedings of the IEEE/CVF Winter Conference on Applications of Computer Vision},
  pages={2920--2930},
  year={2024}
}

@article{he2024dreamstory,
  title={DreamStory: Open-Domain Story Visualization by LLM-Guided Multi-Subject Consistent Diffusion},
  author={He, Huiguo and Yang, Huan and Tuo, Zixi and Zhou, Yuan and Wang, Qiuyue and Zhang, Yuhang and Liu, Zeyu and Huang, Wenhao and Chao, Hongyang and Yin, Jian},
  journal={arXiv preprint arXiv:2407.12899},
  year={2024}
}

@article{zhu2023cogcartoon,
  title={CogCartoon: Towards Practical Story Visualization},
  author={Zhu, Zhongyang and Tang, Jie},
  journal={arXiv preprint arXiv:2312.10718},
  year={2023}
}

@article{jing2015content,
  title={Content-aware video2comics with manga-style layout},
  author={Jing, Guangmei and Hu, Yongtao and Guo, Yanwen and Yu, Yizhou and Wang, Wenping},
  journal={IEEE Transactions on Multimedia},
  volume={17},
  number={12},
  pages={2122--2133},
  year={2015},
  publisher={IEEE}
}

@article{yang2021automatic,
  title={Automatic comic generation with stylistic multi-page layouts and emotion-driven text balloon generation},
  author={Yang, Xin and Ma, Zongliang and Yu, Letian and Cao, Ying and Yin, Baocai and Wei, Xiaopeng and Zhang, Qiang and Lau, Rynson WH},
  journal={ACM Transactions on Multimedia Computing, Communications, and Applications (TOMM)},
  volume={17},
  number={2},
  pages={1--19},
  year={2021},
  publisher={ACM New York, NY, USA}
}

@article{cao2012automatic,
  title={Automatic stylistic manga layout},
  author={Cao, Ying and Chan, Antoni B and Lau, Rynson WH},
  journal={ACM Transactions on Graphics (TOG)},
  volume={31},
  number={6},
  pages={1--10},
  year={2012},
  publisher={ACM New York, NY, USA}
}

@article{proven2021comicgan,
  title={ComicGAN: Text-to-comic generative adversarial network},
  author={Proven-Bessel, Ben and Zhao, Zilong and Chen, Lydia},
  journal={arXiv preprint arXiv:2109.09120},
  year={2021}
}

@inproceedings{luddecke2022image,
  title={Image segmentation using text and image prompts},
  author={L{\"u}ddecke, Timo and Ecker, Alexander},
  booktitle={Proceedings of the IEEE/CVF conference on computer vision and pattern recognition},
  pages={7086--7096},
  year={2022}
}

@inproceedings{gupta2018imagine,
  title={Imagine this! scripts to compositions to videos},
  author={Gupta, Tanmay and Schwenk, Dustin and Farhadi, Ali and Hoiem, Derek and Kembhavi, Aniruddha},
  booktitle={Proceedings of the European conference on computer vision (ECCV)},
  pages={598--613},
  year={2018}
}

@inproceedings{li2019storygan,
  title={Storygan: A sequential conditional gan for story visualization},
  author={Li, Yitong and Gan, Zhe and Shen, Yelong and Liu, Jingjing and Cheng, Yu and Wu, Yuexin and Carin, Lawrence and Carlson, David and Gao, Jianfeng},
  booktitle={Proceedings of the IEEE/CVF conference on computer vision and pattern recognition},
  pages={6329--6338},
  year={2019}
}

@article{maharana2021integrating,
  title={Integrating visuospatial, linguistic and commonsense structure into story visualization},
  author={Maharana, Adyasha and Bansal, Mohit},
  journal={arXiv preprint arXiv:2110.10834},
  year={2021}
}

@inproceedings{maharana2022storydall,
  title={Storydall-e: Adapting pretrained text-to-image transformers for story continuation},
  author={Maharana, Adyasha and Hannan, Darryl and Bansal, Mohit},
  booktitle={European Conference on Computer Vision},
  pages={70--87},
  year={2022},
  organization={Springer}
}

@article{ye2023ip,
  title={Ip-adapter: Text compatible image prompt adapter for text-to-image diffusion models},
  author={Ye, Hu and Zhang, Jun and Liu, Sibo and Han, Xiao and Yang, Wei},
  journal={arXiv preprint arXiv:2308.06721},
  year={2023}
}

@inproceedings{ruiz2023dreambooth,
  title={Dreambooth: Fine tuning text-to-image diffusion models for subject-driven generation},
  author={Ruiz, Nataniel and Li, Yuanzhen and Jampani, Varun and Pritch, Yael and Rubinstein, Michael and Aberman, Kfir},
  booktitle={Proceedings of the IEEE/CVF conference on computer vision and pattern recognition},
  pages={22500--22510},
  year={2023}
}

@article{gal2022image,
  title={An image is worth one word: Personalizing text-to-image generation using textual inversion},
  author={Gal, Rinon and Alaluf, Yuval and Atzmon, Yuval and Patashnik, Or and Bermano, Amit H and Chechik, Gal and Cohen-Or, Daniel},
  journal={arXiv preprint arXiv:2208.01618},
  year={2022}
}

@inproceedings{li2024photomaker,
  title={Photomaker: Customizing realistic human photos via stacked id embedding},
  author={Li, Zhen and Cao, Mingdeng and Wang, Xintao and Qi, Zhongang and Cheng, Ming-Ming and Shan, Ying},
  booktitle={Proceedings of the IEEE/CVF Conference on Computer Vision and Pattern Recognition},
  pages={8640--8650},
  year={2024}
}

@article{li2024blip,
  title={Blip-diffusion: Pre-trained subject representation for controllable text-to-image generation and editing},
  author={Li, Dongxu and Li, Junnan and Hoi, Steven},
  journal={Advances in Neural Information Processing Systems},
  volume={36},
  year={2024}
}

@inproceedings{wei2023elite,
  title={Elite: Encoding visual concepts into textual embeddings for customized text-to-image generation},
  author={Wei, Yuxiang and Zhang, Yabo and Ji, Zhilong and Bai, Jinfeng and Zhang, Lei and Zuo, Wangmeng},
  booktitle={Proceedings of the IEEE/CVF International Conference on Computer Vision},
  pages={15943--15953},
  year={2023}
}

@inproceedings{talecrafter2024,
author = {Gong, Yuan and Pang, Youxin and Cun, Xiaodong and Xia, Menghan and He, Yingqing and Chen, Haoxin and Wang, Longyue and Zhang, Yong and Wang, Xintao and Shan, Ying and Yang, Yujiu},
title = {Interactive Story Visualization with Multiple Characters},
year = {2023},
isbn = {9798400703157},
publisher = {Association for Computing Machinery},
address = {New York, NY, USA},
url = {https://doi.org/10.1145/3610548.3618184},
doi = {10.1145/3610548.3618184},
booktitle = {SIGGRAPH Asia 2023 Conference Papers},
articleno = {101},
numpages = {10},
location = {Sydney, NSW, Australia},
series = {SA '23}
}

@article{wang2023autostory,
  title={AutoStory: Generating Diverse Storytelling Images with Minimal Human Efforts},
  author={Wang, Wen and Zhao, Canyu and Chen, Hao and Chen, Zhekai and Zheng, Kecheng and Shen, Chunhua},
  journal={International Journal of Computer Vision},
  pages={1--22},
  year={2024},
  publisher={Springer}
}

@article{wang2024large,
  title={Large language models are latent variable models: Explaining and finding good demonstrations for in-context learning},
  author={Wang, Xinyi and Zhu, Wanrong and Saxon, Michael and Steyvers, Mark and Wang, William Yang},
  journal={Advances in Neural Information Processing Systems},
  volume={36},
  year={2024}
}

@article{zhang2023multimodal,
  title={Multimodal chain-of-thought reasoning in language models},
  author={Zhang, Zhuosheng and Zhang, Aston and Li, Mu and Zhao, Hai and Karypis, George and Smola, Alex},
  journal={arXiv preprint arXiv:2302.00923},
  year={2023}
}

@article{nguyen2018digital,
  title={Digital comics image indexing based on deep learning},
  author={Nguyen, Nhu-Van and Rigaud, Christophe and Burie, Jean-Christophe},
  journal={Journal of Imaging},
  volume={4},
  number={7},
  pages={89},
  year={2018},
  publisher={MDPI}
}

@article{achiam2023gpt,
  title={Gpt-4 technical report},
  author={Achiam, Josh and Adler, Steven and Agarwal, Sandhini and Ahmad, Lama and Akkaya, Ilge and Aleman, Florencia Leoni and Almeida, Diogo and Altenschmidt, Janko and Altman, Sam and Anadkat, Shyamal and others},
  journal={arXiv preprint arXiv:2303.08774},
  year={2023}
}

@article{gu2024mix,
  title={Mix-of-show: Decentralized low-rank adaptation for multi-concept customization of diffusion models},
  author={Gu, Yuchao and Wang, Xintao and Wu, Jay Zhangjie and Shi, Yujun and Chen, Yunpeng and Fan, Zihan and Xiao, Wuyou and Zhao, Rui and Chang, Shuning and Wu, Weijia and others},
  journal={Advances in Neural Information Processing Systems},
  volume={36},
  year={2024}
}

@article{tang2022daam,
  title={What the daam: Interpreting stable diffusion using cross attention},
  author={Tang, Raphael and Liu, Linqing and Pandey, Akshat and Jiang, Zhiying and Yang, Gefei and Kumar, Karun and Stenetorp, Pontus and Lin, Jimmy and Ture, Ferhan},
  journal={arXiv preprint arXiv:2210.04885},
  year={2022}
}

@article{heusel2017gans,
  title={Gans trained by a two time-scale update rule converge to a local nash equilibrium},
  author={Heusel, Martin and Ramsauer, Hubert and Unterthiner, Thomas and Nessler, Bernhard and Hochreiter, Sepp},
  journal={Advances in neural information processing systems},
  volume={30},
  year={2017}
}

@inproceedings{hore2010image,
  title={Image quality metrics: PSNR vs. SSIM},
  author={Hore, Alain and Ziou, Djemel},
  booktitle={2010 20th international conference on pattern recognition},
  pages={2366--2369},
  year={2010},
  organization={IEEE}
}

@inproceedings{li2022blip,
  title={Blip: Bootstrapping language-image pre-training for unified vision-language understanding and generation},
  author={Li, Junnan and Li, Dongxu and Xiong, Caiming and Hoi, Steven},
  booktitle={International Conference on Machine Learning},
  pages={12888--12900},
  year={2022},
  organization={PMLR}
}

@misc{Aes2022,
  title={Improved aesthetic predictor},
  author={Christoph Schuhmann},
  url={https://github.com/christophschuhmann/improved-aesthetic-predictor},
  year={2022},
}

@inproceedings{wu2023human,
  title={Human preference score: Better aligning text-to-image models with human preference},
  author={Wu, Xiaoshi and Sun, Keqiang and Zhu, Feng and Zhao, Rui and Li, Hongsheng},
  booktitle={Proceedings of the IEEE/CVF International Conference on Computer Vision},
  pages={2096--2105},
  year={2023}
}

@article{vivoli2024comix,
  title={CoMix: A Comprehensive Benchmark for Multi-Task Comic Understanding},
  author={Vivoli, Emanuele and Bertini, Marco and Karatzas, Dimosthenis},
  journal={arXiv preprint arXiv:2407.03550},
  year={2024}
}

@article{touvron2023llama,
  title={Llama 2: Open foundation and fine-tuned chat models},
  author={Touvron, Hugo and Martin, Louis and Stone, Kevin and Albert, Peter and Almahairi, Amjad and Babaei, Yasmine and Bashlykov, Nikolay and Batra, Soumya and Bhargava, Prajjwal and Bhosale, Shruti and others},
  journal={arXiv preprint arXiv:2307.09288},
  year={2023}
}

@article{chiang2023vicuna,
  title={Vicuna: An open-source chatbot impressing gpt-4 with 90\%* chatgpt quality},
  author={Chiang, Wei-Lin and Li, Zhuohan and Lin, Zi and Sheng, Ying and Wu, Zhanghao and Zhang, Hao and Zheng, Lianmin and Zhuang, Siyuan and Zhuang, Yonghao and Gonzalez, Joseph E and others},
  journal={See https://vicuna. lmsys. org (accessed 14 April 2023)},
  volume={2},
  number={3},
  pages={6},
  year={2023}
}

@inproceedings{gatys2017controlling,
  title={Controlling perceptual factors in neural style transfer},
  author={Gatys, Leon A and Ecker, Alexander S and Bethge, Matthias and Hertzmann, Aaron and Shechtman, Eli},
  booktitle={Proceedings of the IEEE conference on computer vision and pattern recognition},
  pages={3985--3993},
  year={2017}
}

@inproceedings{hong2018inferring,
  title={Inferring semantic layout for hierarchical text-to-image synthesis},
  author={Hong, Seunghoon and Yang, Dingdong and Choi, Jongwook and Lee, Honglak},
  booktitle={Proceedings of the IEEE conference on computer vision and pattern recognition},
  pages={7986--7994},
  year={2018}
}

@inproceedings{yin2017obj2text,
  title={Obj2Text: Generating Visually Descriptive Language from Object Layouts},
  author={Yin, Xuwang and Ordonez, Vicente},
  booktitle={Proceedings of the 2017 Conference on Empirical Methods in Natural Language Processing},
  pages={177--187},
  year={2017}
}

@inproceedings{banerjee2005meteor,
  title={METEOR: An automatic metric for MT evaluation with improved correlation with human judgments},
  author={Banerjee, Satanjeev and Lavie, Alon},
  booktitle={Proceedings of the acl workshop on intrinsic and extrinsic evaluation measures for machine translation and/or summarization},
  pages={65--72},
  year={2005}
}

@inproceedings{lin2004rouge,
  title={Rouge: A package for automatic evaluation of summaries},
  author={Lin, Chin-Yew},
  booktitle={Text summarization branches out},
  pages={74--81},
  year={2004}
}

@inproceedings{vedantam2015cider,
  title={Cider: Consensus-based image description evaluation},
  author={Vedantam, Ramakrishna and Lawrence Zitnick, C and Parikh, Devi},
  booktitle={Proceedings of the IEEE conference on computer vision and pattern recognition},
  pages={4566--4575},
  year={2015}
}

@inproceedings{anderson2016spice,
  title={Spice: Semantic propositional image caption evaluation},
  author={Anderson, Peter and Fernando, Basura and Johnson, Mark and Gould, Stephen},
  booktitle={Computer Vision--ECCV 2016: 14th European Conference, Amsterdam, The Netherlands, October 11-14, 2016, Proceedings, Part V 14},
  pages={382--398},
  year={2016},
  organization={Springer}
}

@inproceedings{tan2019text2scene,
  title={Text2scene: Generating compositional scenes from textual descriptions},
  author={Tan, Fuwen and Feng, Song and Ordonez, Vicente},
  booktitle={Proceedings of the IEEE/CVF Conference on Computer Vision and Pattern Recognition},
  pages={6710--6719},
  year={2019}
}

@inproceedings{lin2025evaluating,
  title={Evaluating text-to-visual generation with image-to-text generation},
  author={Lin, Zhiqiu and Pathak, Deepak and Li, Baiqi and Li, Jiayao and Xia, Xide and Neubig, Graham and Zhang, Pengchuan and Ramanan, Deva},
  booktitle={European Conference on Computer Vision},
  pages={366--384},
  year={2025},
  organization={Springer}
}

@inproceedings{guerin2013ebdtheque,
  title={eBDtheque: a representative database of comics},
  author={Gu{\'e}rin, Cl{\'e}ment and Rigaud, Christophe and Mercier, Antoine and Ammar-Boudjelal, Farid and Bertet, Karell and Bouju, Alain and Burie, Jean-Christophe and Louis, Georges and Ogier, Jean-Marc and Revel, Arnaud},
  booktitle={2013 12th International Conference on Document Analysis and Recognition},
  pages={1145--1149},
  year={2013},
  organization={IEEE}
}

@article{organisciak2023beyond,
  title={Beyond semantic distance: Automated scoring of divergent thinking greatly improves with large language models},
  author={Organisciak, Peter and Acar, Selcuk and Dumas, Denis and Berthiaume, Kelly},
  journal={Thinking Skills and Creativity},
  volume={49},
  pages={101356},
  year={2023},
  publisher={Elsevier}
}

@article{dong2024iclsurvey,
    title={A Survey on In-context Learning}, 
    author={Qingxiu Dong and Lei Li and Damai Dai and Ce Zheng and Jingyuan Ma and Rui Li and Heming Xia and Jingjing Xu and Zhiyong Wu and Tianyu Liu and Baobao Chang and Xu Sun and Lei Li and Zhifang Sui},
    year={2024},
    journal={arXiv}
}

@article{paszke2019pytorch,
  title={Pytorch: An imperative style, high-performance deep learning library},
  author={Paszke, Adam and Gross, Sam and Massa, Francisco and Lerer, Adam and Bradbury, James and Chanan, Gregory and Killeen, Trevor and Lin, Zeming and Gimelshein, Natalia and Antiga, Luca and others},
  journal={Advances in neural information processing systems},
  volume={32},
  year={2019}
}

@inproceedings{li2023gligen,
  title={Gligen: Open-set grounded text-to-image generation},
  author={Li, Yuheng and Liu, Haotian and Wu, Qingyang and Mu, Fangzhou and Yang, Jianwei and Gao, Jianfeng and Li, Chunyuan and Lee, Yong Jae},
  booktitle={Proceedings of the IEEE/CVF Conference on Computer Vision and Pattern Recognition},
  pages={22511--22521},
  year={2023}
}

@misc{huggingfaceXenovagpt35turboHugging,
    author = {},
    title = {{X}enova/gpt-3.5-turbo · {H}ugging {F}ace --- huggingface.co},
    howpublished = {\url{https://huggingface.co/Xenova/gpt-3.5-turbo}},
    year = {2024},
    note = {[Accessed 12-11-2024]},
}

@misc{huggingfaceTheBlokeclaude2alpaca13BGPTQHugging,
  author = {Lichang Chen and Khalid Saifullah and Ming Li and Tianyi Zhou and Heng Huang},
  title = {Claude2-Alpaca: Instruction tuning datasets distilled from claude},
  year = {2023},
  publisher = {GitHub},
  journal = {GitHub repository},
  howpublished = {\url{https://github.com/Lichang-Chen/claude2-alpaca}},
}

@misc{anthropicClaudeSonnet,
	author = {},
	title = {{C}laude 3.5 {S}onnet --- anthropic.com},
	howpublished = {\url{https://www.anthropic.com/claude/sonnet}},
	year = {},
	note = {[Accessed 12-11-2024]},
}

@article{githubGitHubHLasseTextDescriptives,
   title={TextDescriptives: A Python package for calculating a
large variety of metrics from text},
   volume={8},
   ISSN={2475-9066},
   url={http://dx.doi.org/10.21105/joss.05153},
   DOI={10.21105/joss.05153},
   number={84},
   journal={Journal of Open Source Software},
   publisher={The Open Journal},
   author={Hansen, Lasse and Olsen, Ludvig Renbo and Enevoldsen, Kenneth},
   year={2023},
   month=apr, pages={5153} }

@misc{huggingfaceBlackforestlabsFLUX1schnellHugging,
    author={Black Forest Labs},
    title={FLUX},
    year={2024},
    howpublished={\url{https://github.com/black-forest-labs/flux}},
}

@misc{huggingfaceRenderartistretrocomicfluxHugging,
	author = {},
	title = {renderartist/retrocomicflux · {H}ugging {F}ace --- huggingface.co},
	howpublished = {\url{https://huggingface.co/renderartist/retrocomicflux}},
	year = {2024},
	note = {[Accessed 12-11-2024]},
}

@article{dahary2024yourself,
  title={Be yourself: Bounded attention for multi-subject text-to-image generation},
  author={Dahary, Omer and Patashnik, Or and Aberman, Kfir and Cohen-Or, Daniel},
  journal={arXiv preprint arXiv:2403.16990},
  volume={2},
  number={5},
  year={2024}
}

@inproceedings{liu2024improved,
  title={Improved baselines with visual instruction tuning},
  author={Liu, Haotian and Li, Chunyuan and Li, Yuheng and Lee, Yong Jae},
  booktitle={Proceedings of the IEEE/CVF Conference on Computer Vision and Pattern Recognition},
  pages={26296--26306},
  year={2024}
}

@article{cao2024synartifact,
  title={SynArtifact: Classifying and Alleviating Artifacts in Synthetic Images via Vision-Language Model},
  author={Cao, Bin and Yuan, Jianhao and Liu, Yexin and Li, Jian and Sun, Shuyang and Liu, Jing and Zhao, Bo},
  journal={arXiv preprint arXiv:2402.18068},
  year={2024}
}

@inproceedings{wang2023cdac,
  title={CDAC: Cross-domain attention consistency in transformer for domain adaptive semantic segmentation},
  author={Wang, Kaihong and Kim, Donghyun and Feris, Rogerio and Betke, Margrit},
  booktitle={Proceedings of the IEEE/CVF International Conference on Computer Vision},
  pages={11519--11529},
  year={2023}
}

@inproceedings{liuone,
  title={One-Prompt-One-Story: Free-Lunch Consistent Text-to-Image Generation Using a Single Prompt},
  author={Liu, Tao and Wang, Kai and Li, Senmao and van de Weijer, Joost and Khan, Fahad Shahbaz and Yang, Shiqi and Wang, Yaxing and Yang, Jian and Cheng, Ming-Ming},
  booktitle={The Thirteenth International Conference on Learning Representations},
  year = {2025}
}

@article{wu2024diffsensei,
  title={DiffSensei: Bridging Multi-Modal LLMs and Diffusion Models for Customized Manga Generation},
  author={Wu, Jianzong and Tang, Chao and Wang, Jingbo and Zeng, Yanhong and Li, Xiangtai and Tong, Yunhai},
  journal={arXiv preprint arXiv:2412.07589},
  year={2024}
}

@article{singh2025storybooth,
  title={Storybooth: Training-free Multi-Subject Consistency for Improved Visual Storytelling},
  author={Singh, Jaskirat and Chen, Junshen Kevin and Kohler, Jonas and Cohen, Michael},
  journal={arXiv preprint arXiv:2504.05800},
  year={2025}
}

@inproceedings{chen2024mangadiffusion,
    title={Manga Generation via Layout-controllable Diffusion},
    author={Chen, Siyu and Li, Dengjie and Bao, Zenghao and Zhou, Yao and Tan, Lingfeng and Zhong, Yujie and Zhao, Zheng},
    booktitle={arXiv preprint arxiv:2412.19303},
    year={2024}
}

@article{li2024unbounded,
  title={Unbounded: A Generative Infinite Game of Character Life Simulation},
  author={Li, Jialu and Li, Yuanzhen and Wadhwa, Neal and Pritch, Yael and Jacobs, David E and Rubinstein, Michael and Bansal, Mohit and Ruiz, Nataniel},
  journal={arXiv preprint arXiv:2410.18975},
  year={2024}
}

@article{he2024improving,
  title={Improving Multi-Subject Consistency in Open-Domain Image Generation with Isolation and Reposition Attention},
  author={He, Huiguo and Wang, Qiuyue and Zhou, Yuan and Cai, Yuxuan and Chao, Hongyang and Yin, Jian and Yang, Huan},
  journal={arXiv preprint arXiv:2411.19261},
  year={2024}
}

@article{jang2024identity,
  title={Identity decoupling for multi-subject personalization of text-to-image models},
  author={Jang, Sangwon and Jo, Jaehyeong and Lee, Kimin and Hwang, Sung Ju},
  journal={arXiv preprint arXiv:2404.04243},
  year={2024}
}

@article{geyer2023tokenflow,
  title={Tokenflow: Consistent diffusion features for consistent video editing},
  author={Geyer, Michal and Bar-Tal, Omer and Bagon, Shai and Dekel, Tali},
  journal={arXiv preprint arXiv:2307.10373},
  year={2023}
}

@inproceedings{hu2024animate,
  title={Animate anyone: Consistent and controllable image-to-video synthesis for character animation},
  author={Hu, Li},
  booktitle={Proceedings of the IEEE/CVF Conference on Computer Vision and Pattern Recognition},
  pages={8153--8163},
  year={2024}
}
}

\clearpage
\setcounter{page}{1}
\twocolumn[{%
\renewcommand\twocolumn[1][]{#1}%
\maketitlesupplementary
\begin{center}
    \centering
    \captionsetup{type=figure}
    \includegraphics[width=\textwidth]{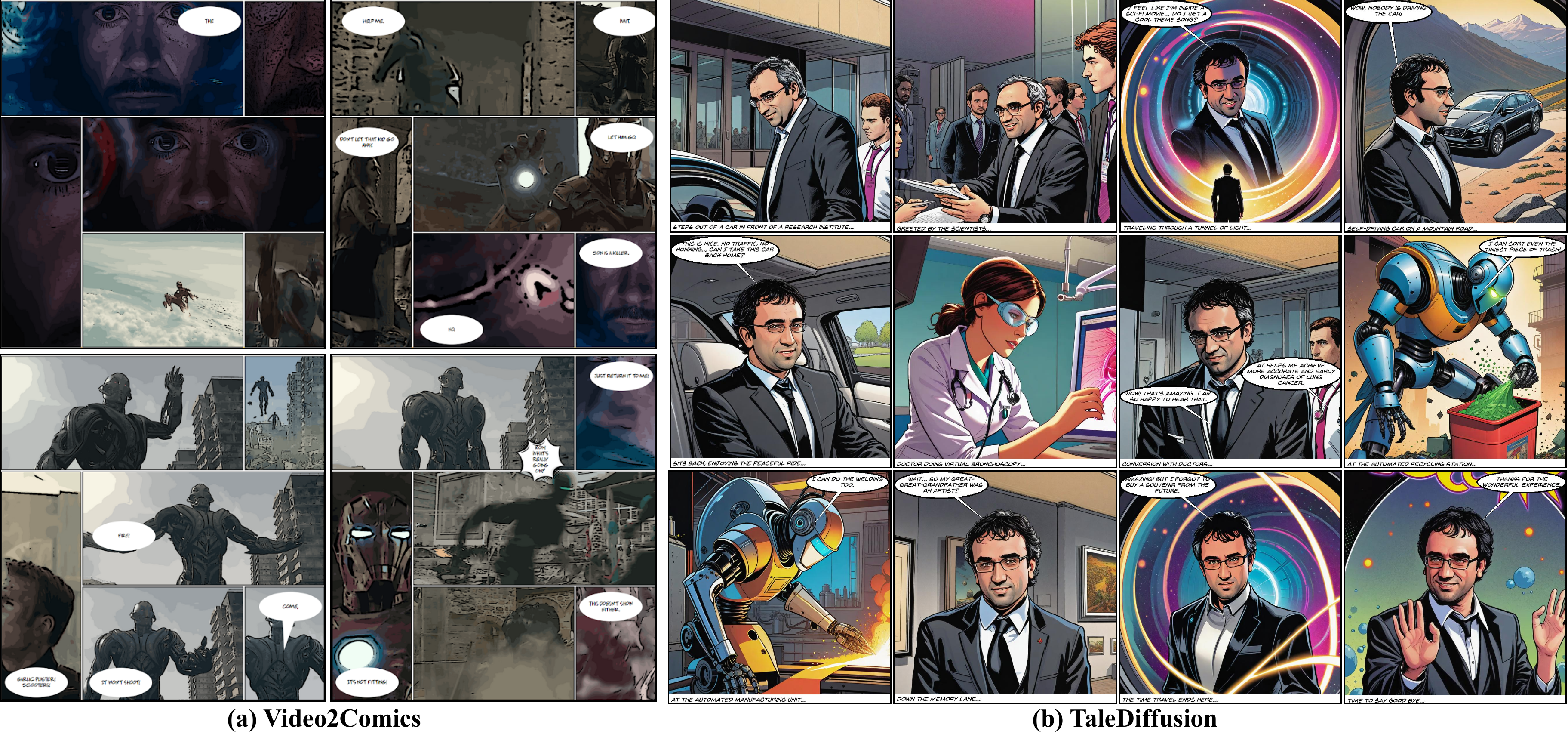}
    \captionof{figure}{\textbf{Video2Comics \cite{yang2021automatic} vs TaleDiffusion:} The former struggles with character consistency and dialogue accuracy, while the latter ensures both, maintaining story coherence [500\% zoom].}
    \label{fig:vidcomp}
\end{center}%
}]

\section{Implementation Details}
\label{sec:imp}
This section offers in-depth insights into the implementation of TaleDiffusion as a training-free framework that is compatible with most existing
LLM architectures and diffusion models. We implemented our model using PyTorch \cite{paszke2019pytorch} and leveraged the Diffusion models from diffuser \footnote[1]{https://github.com/huggingface/diffusers}. We have utilized the GPT4 \cite{achiam2023gpt} model for story expansion and layout generation as it provides satisfactory performance for both tasks. 

Similarly, we use the diffusion U-Net of GLIGEN \cite{li2023gligen} without further training and adaptation. This selected GLIGEN model used stable diffusion v1.5 from the diffusers with additional layout guidance. The guidance scale is set to  7.5 as mentioned in \cite{li2023gligen}. For the energy minimization in image synthesis, we repeat the optimization step 5 times for every object denoising step until the repetition is reduced to 1. Also, $\xi$ is set to 10 as a sharpness, and the $\phi_j$ is set to 0.2 as a soft threshold of mask generation.

For consistent character generation, we use IP-Adapter \cite{ye2023ip} with a subject guidance factor ($\lambda$) of 0.95 and an image intensity factor of 0.7. We finetune this IP-Adapter for a multi-character customization process with low-rank adaptation with rank ($\sigma$) 64 and regularization factor ($\alpha$) 128 on a single GPU of NVIDIA A100. We run the denoising process with 100 steps by default. We only perform latent compose in the first quarter of the denoising process (first 25 steps). For the bubble assignment, we utilize the CLIPSeg weights taken from its official repository \footnote[2]{https://huggingface.co/CIDAS/clipseg-rd64-refined}. 

\section{Chain-of-thoughts vs In-Context learning}
\label{sec:cot}
With Chain-of-thoughts \cite{he2024dreamstory} story expansion we guide the model through a series of interconnected prompts as depicted in \cref{fig:LLMs}. In Chain-of-thoughts \cite{he2024dreamstory} prompting each prompt builds on the previous one, creating a continuous “chain” of thoughts or ideas. In order to expand a story from the short description $S_{con}$ it first generates the set of panel description $\mathcal{B}$. Based on the generated panel description they predicted the set of character $\mathcal{C}$ and subsequently predicted the set of dialogues $\mathfrak{D}$ based on $\mathcal{C}$. Based on this fact, the entropy function of the Chain-of-thoughts \cite{he2024dreamstory} can be designed as defined in \cref{eq:supp1}.
\begin{equation}
    \label{eq:supp1}
    \mathcal{B}, \mathcal{C}, \mathfrak{D} = \argmax_{\mathcal{B}_j, \mathfrak{D}_j} \sum_{j=1}^k P(B_j|S_{con})\text{ln} P(C_j|B_j) \text{ln} P(D_j|C_j)
\end{equation}

Where, $k$ is the number of iterations to optimize the $\mathcal{B}, \mathcal{C}, \mathfrak{D}$. Although Chain-of-thoughts \cite{he2024dreamstory} provides consistent and reliable outcomes in question answering by preserving context and linking responses keeps the model focused on the topic or problem, it is not the convenient method for story expansion as depicted in \cref{tab:04}. As the entropy function of Chain-of-thoughts \cite{he2024dreamstory} depends on a set of conditional probability functions one bad prediction can affect the performance of the rest and character prediction ($\mathcal{C}$) from the panel description ($\mathcal{B}$) always leads to poor performance due to lack of clarity of the character descriptions (i.e. detailing of dressing, face, hairstyle and so on) in $\mathcal{B}$.

From the \cref{tab:04}, it can be concluded that in-context learning beat the performance of Chain-of-thoughts for all the LLMs ($\approx$ 8\%) in readability, coherence, and creativity and creativity as in-context learning capitalizes on the model’s inherent capacity to generalize and adapt to new situations based on input received during user interactions.

\section{Additional details on layout generation}

\begin{algorithm}[t]
\begin{algorithmic}
\Require{User prompt $T_p$, Initial layout $L_i$} 
\Ensure{Final layout $L_f$.}
\State set $i \gets 1$;\\
\State set $\delta_{rec}^{i-1} \gets 1$;\\
\State set $\delta_{rec}^i \gets 10e9+7$;\\
\While{$\delta_{rec} \neq 0$ and $|\delta_{rec}^{i-1}-\delta_{rec}^i|<1e-4$}
    \State reconstruct caption with \cite{yin2017obj2text};\\
    \State update $\delta_{rec}^i$ as defined in \cref{sec:m1};\\
    \State set $\delta_{rec}^{i-1} \gets \delta_{rec}$;\\
    \State asks the LLM to reconstruct the layout with updated $\delta_{rec}$;\\
    \State $i \gets i+1$;
\EndWhile
\end{algorithmic}
\caption{Iterative layout correction}
\label{algo1}
\end{algorithm}

This study aims to identify the optimal LLM for generating accurate layouts from text prompts that convey enumeration, position, relationships, and spatial arrangements of multiple objects. Following \cite{hong2018inferring}, we use caption generation as an extrinsic evaluation method, employing \cite{yin2017obj2text} to generate captions from semantic layouts for comparison with original scene captions (see \cref{algo1} for more details). We assess performance using existing metrics like METEOR \cite{banerjee2005meteor}, ROUGE \cite{lin2004rouge}, CIDEr \cite{vedantam2015cider}, and SPICE \cite{anderson2016spice}, as detailed in \cref{tab:05}. The performances are based on 50 unique prompts tested 5 times each, without layout correction, to determine the best LLM for our work. We started testing the Text2Scene module \cite{tan2019text2scene}, which struggled with objects outside the MS-COCO dataset. We then explored LLMs, starting with Llama-7B \cite{touvron2023llama}, which improved the baseline by 8\%. Vicuna-13B \cite{chiang2023vicuna} further refined Llama-7B, offering a 5\% improvement, and GPT-3.5 surpassed Vicuna-13B by 3\%. Between GPT4o and Claude-3.5 Sonnet, with a 2:3 ratio favoring Claude-3.5 Sonnet, we chose the former for our experiments due to compatibility (as in story expansion, GPT4o \cite{achiam2023gpt} provides the superior performance, and layout generation is a dependent task on story expansion).

\begin{table*}[!htbp]
\centering
\caption{Evaluation of layout generation with LLMs.}
\resizebox{\textwidth}{!}{
\begin{tabular}{c|ccccc|ccccc}
\hline
\multirow{2}{*}{Methods/Metrics} & \multicolumn{5}{c|}{w/o reconstruction error} & \multicolumn{5}{c}{With reconstruction error} \\ \cline{2-11} 
 &
  \multicolumn{1}{c|}{METEOR \cite{banerjee2005meteor} $\uparrow$} &
  \multicolumn{1}{c|}{ROUGE \cite{lin2004rouge} $\uparrow$} &
  \multicolumn{1}{c|}{CIDEr \cite{vedantam2015cider} $\uparrow$} &
  \multicolumn{1}{c|}{SPICE \cite{anderson2016spice} $\uparrow$} &
  Average $\uparrow$ &
  \multicolumn{1}{c|}{METEOR \cite{banerjee2005meteor} $\uparrow$} &
  \multicolumn{1}{c|}{ROUGE \cite{lin2004rouge} $\uparrow$} &
  \multicolumn{1}{c|}{CIDEr \cite{vedantam2015cider} $\uparrow$} &
  \multicolumn{1}{c|}{SPICE \cite{anderson2016spice} $\uparrow$} &
  Average $\uparrow$ \\ \hline
Text2Scene \cite{tan2019text2scene}                      & 0.168   & 0.397   & 0.572   & 0.103  & 0.310  & 0.189   & 0.446   & 0.601   & 0.123  & 0.339  \\
Llama-7B \cite{touvron2023llama}                         & 0.211   & 0.491   & 0.617   & 0.190  & 0.377  & 0.232   & 0.573   & 0.648   & 0.214  & 0.416  \\
Vicuna-13B \cite{chiang2023vicuna}                       & 0.250   & 0.571   & 0.709   & 0.202  & 0.433  & 0.274   & 0.656   & 0.721   & 0.222  & 0.468  \\
GPT-3.5 \footnote[1]{}                         & 0.262   & 0.682   & 0.694   & 0.176  & 0.453  & 0.271   & 0.711   & 0.753   & 0.254  & 0.497  \\
GPT-4 \cite{achiam2023gpt}                          & 0.271   & 0.715   & 0.702   & 0.342  & 0.507  & 0.289   & 0.737   & 0.797   & 0.382  & 0.511  \\
Claude-2 \footnote[2]{}                         & 0.267   & 0.671   & 0.767   & 0.217  & 0.480  & 0.291   & 0.717   & 0.782   & 0.291  & 0.538  \\
Claude-3.5 Sonnet \footnote[3]{}               & 0.282   & 0.702   & 0.792   & 0.301  & 0.519  & 0.302   & 0.728   & 0.801   & 0.322  & 0.544  \\ \hline
\end{tabular}
}
\label{tab:05}
\end{table*}

\section{Details of Mask Generation}
\label{sec:mask}
\begin{figure*}[!htbp]
\centering
\includegraphics[width=\linewidth]{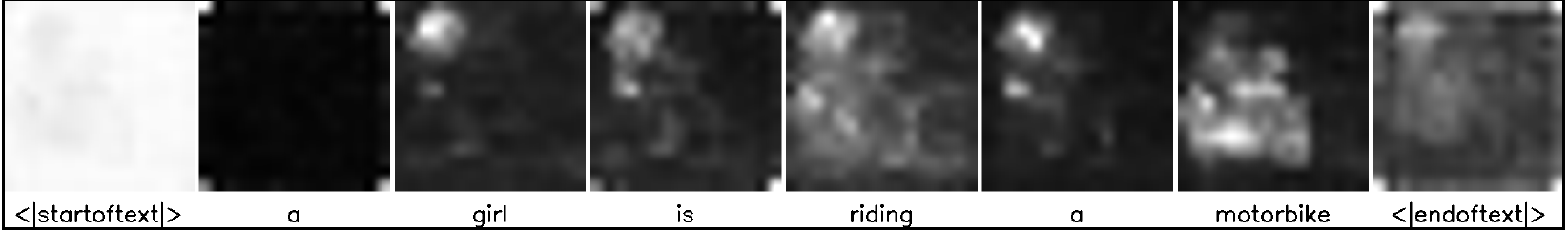}
\caption{Cross attention computation during mask generation}
\label{fig:mask1} 
\end{figure*}

To derive the self-attention mask within a bounding box, we utilized a self-segmentation technique based on the self-attention and cross-attention maps extracted from the diffusion U-Net. In order to do that, we encode the text prompt ($F_{i,j}$) with a CLIP text encoder to get the text tokens and encode the bounding box region ($R_{i,j}$) with a CLIP vision encoder to bring them into the same embedding space. Then we compute the cross attention between them (see \cref{fig:mask1}) as defined in \cref{eq:2}.

In \cref{fig:mask1}, the first and last tokens are the CLIP-generated tokens dedicated to implying the start and end of the foreground prompt. Also, it has been observed that, for each token, it tried to attend to some pixel regions that are suffering from the noise. In order to solve this issue, we extract the self-attention maps from the pretrained diffusion U-Net (specifically from  UNet’s middle block and first up-block of the U-Net are more robust to noise \cite{dahary2024yourself}).

\begin{figure*}[!htbp]
\centering
\includegraphics[width=\linewidth]{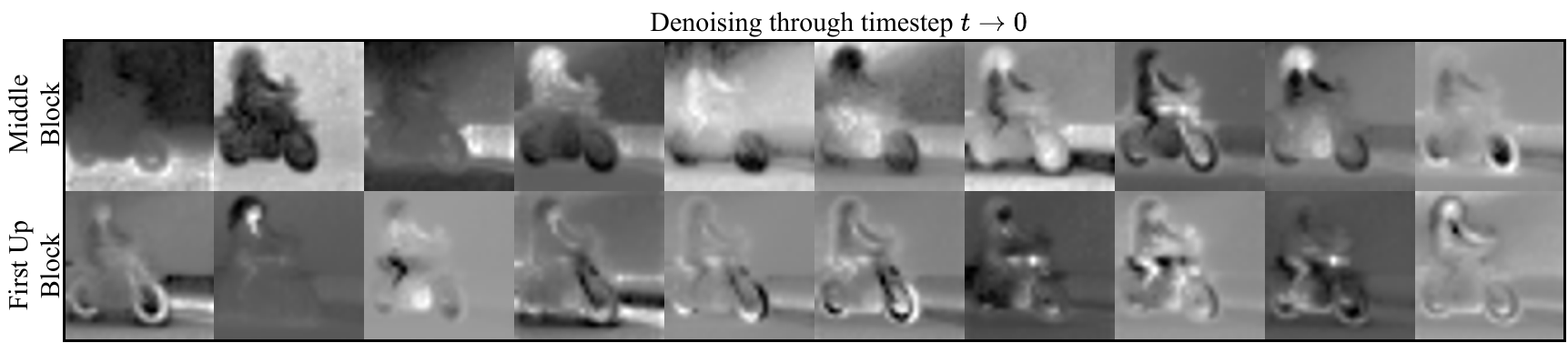}
\caption{Self Attention maps from U-Net's middle and first up block across all the timesteps from t $\rightarrow$ 0.}
\label{fig:mask2} 
\end{figure*}

\begin{figure*}[!htbp]
\centering
\includegraphics[width=\linewidth]{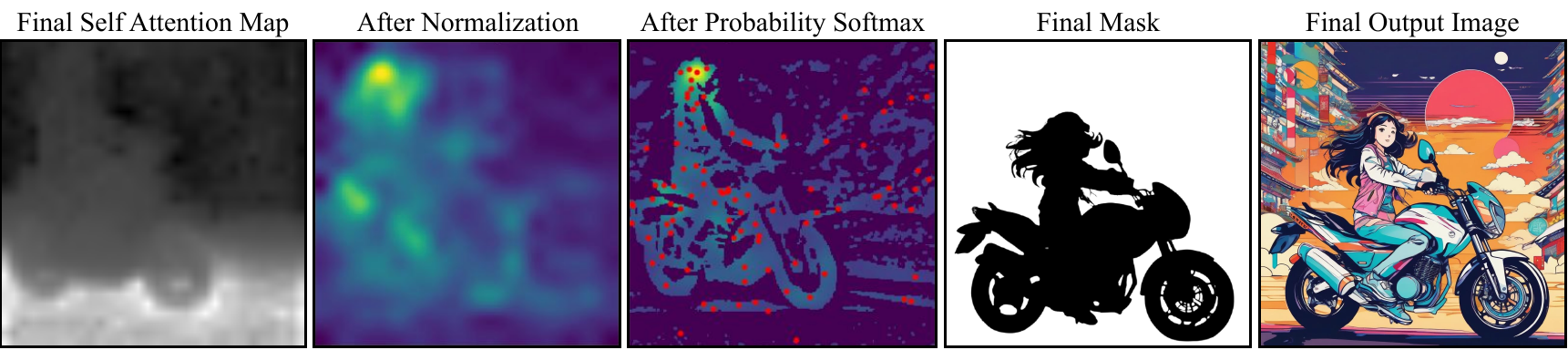}
\caption{Attention map to mask generation}
\label{fig:mask3} 
\end{figure*}

As depicted in \cref{fig:mask2}, these attention maps provide a more prominent shape and texture compared to the previous one in \cref{fig:mask1}. These attention maps are then averaged by these two layers by aggregating the mean of all the timesteps to get the final self-attention map (see \cref{fig:mask3}). 

On top of this self-attention, we apply normalization to highlight some particular region (especially the face, as in the comic face is the most important region). Then we compute the probability softmax function on top of the normalized self-attention to derive the shape and texture of the mask. Last but not least, we apply binarization with coordinate regularization (in order to put the object mask at the center of the bounding box) to get the final object mask corresponding to the bounding box. This process has been performed iteratively in order to generate the complete segmentation mask for the panel from the layout.

\section{Identity Consistent Self-Attention (ICSA)}
\label{sec:RICSA}
\begin{figure*}[!htbp]
\centering
\includegraphics[width=\textwidth]{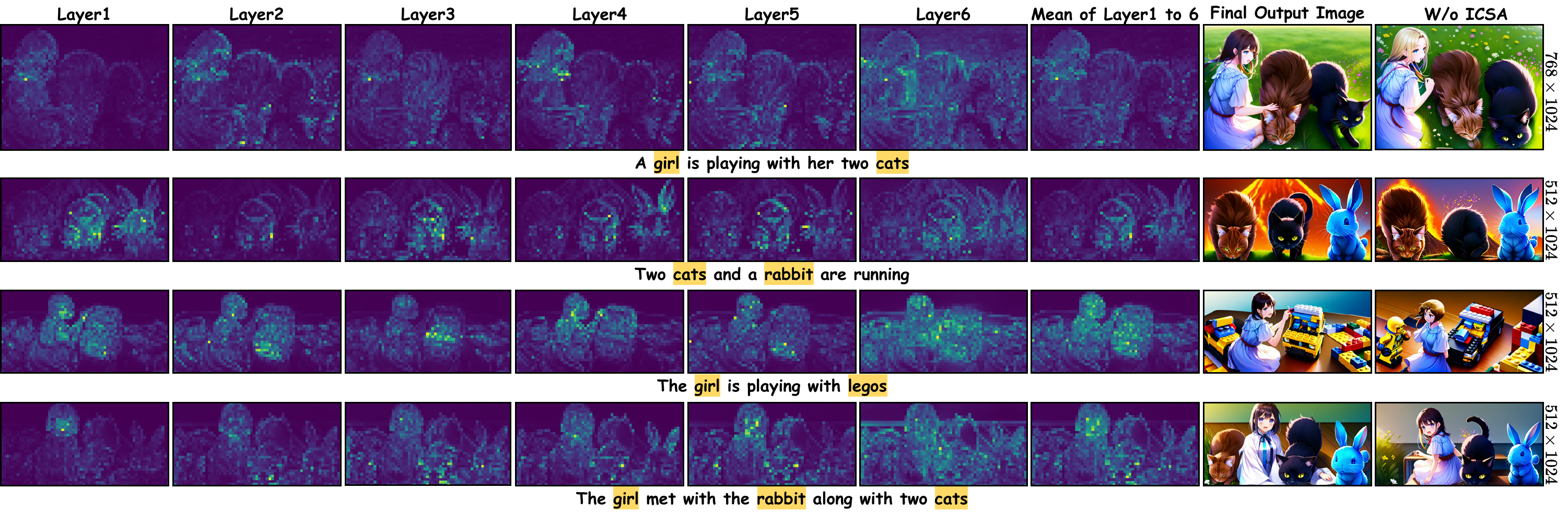}
\vspace{-15pt}
\caption{\textbf{Effectiveness of ICSA:} At the initial layers, ICSA focused on individual characters that maintain character consistency. At a later stage, it attends all the objects together and helps in multi-character customization.}
\label{fig:icsa} 
\end{figure*}

In order to understand how identity-consistent self-attention (ICSA) works, we have visualized the layer-wise ICSA map for a complete story in \cref{fig:icsa}. It has been observed that, in the initial layers, ICSA attends the pixels of the facial regions, and later, they are just aggregating the attended pixels to generate a global self-attention map. Also, it tried to put more focus on human faces than non-human ones, as it is difficult to maintain human facial characteristics. Not only that, we extract the mean ICSA, which helps to maintain consistency during latent denoising of the background generation. We have also tried to generate a story without ICSA in \cref{fig:icsa}. It was observed that, without ICSA, the model struggles to generate the story by maintaining hairstyle, face, pose, and so on, which establishes the effectiveness of ICSA for consistent character generation.

As most of the story generation utilized subject-driven self-attention (SDSA) \cite{tewel2024training} for consistent character generation. In order to, compare the effectiveness of the SDSA and ICSA-RACA combination, we select the same prompt with a single character and visualize the final results as well as the attention maps in \cref{fig:ricsa}. It has been observed the SDSA only attends to the pixels of the eyes and forehead/hair, which restricts them to working up to 4 characters with \cite{tewel2024training}. In contrast, the ICSA helps to maintain the object pose, and the RACA focuses on the overall human face, other objects, and their corresponding positions.

\begin{figure}[!htbp]
\centering
\includegraphics[width=\linewidth]{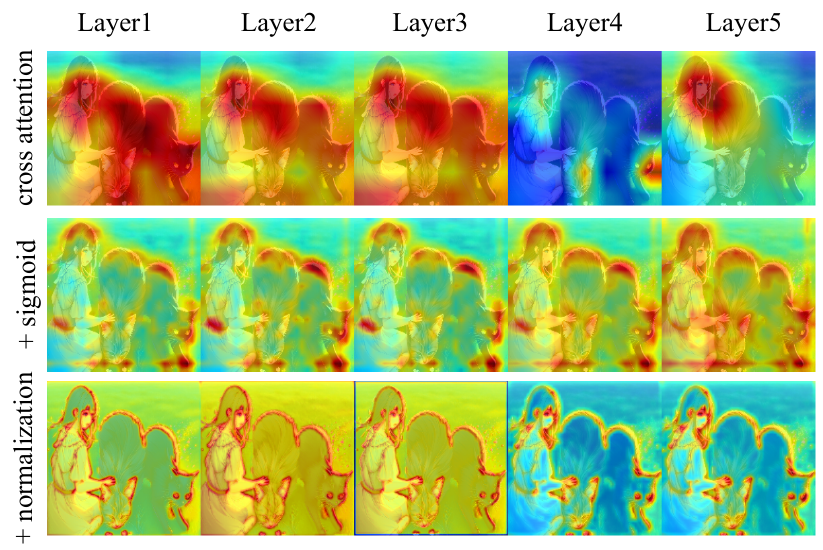}
\caption{\textbf{RACA:} cross-attention, aided by sigmoid normalization, captures object shape and structure for precise positioning.}
\label{fig:cross} 
\end{figure}

\section{Details of Artifacts Score}
\label{sec:artifacts}
A major challenge in the fast-evolving field of image synthesis is addressing complex artifacts that reduce the perceptual realism of synthetic images. However, there is no such metric to evaluate the percentage of artifacts generated by an image synthesis technique. To measure the artifacts score, we first identify the types of artifacts generated by diffusion models and rank them based on their frequency of occurrence. We design a comprehensive artifact taxonomy for synthetic images including 26 kinds of artifacts (bad anatomy, out-of-frame, ugly, duplicate, morbid, multilated, extrafingers, mutated hands, deformed, extra limbs, cloned face, disfigured, malformed limbs, missing arms, missing legs, extra arms, extra legs, fused fingers, too many fingers, long neck and so on) fine tune a vision language model (VLM) to classify and rank them based on frequency of occurrence.

\begin{figure}[!htbp]
\centering
\includegraphics[width=\columnwidth]{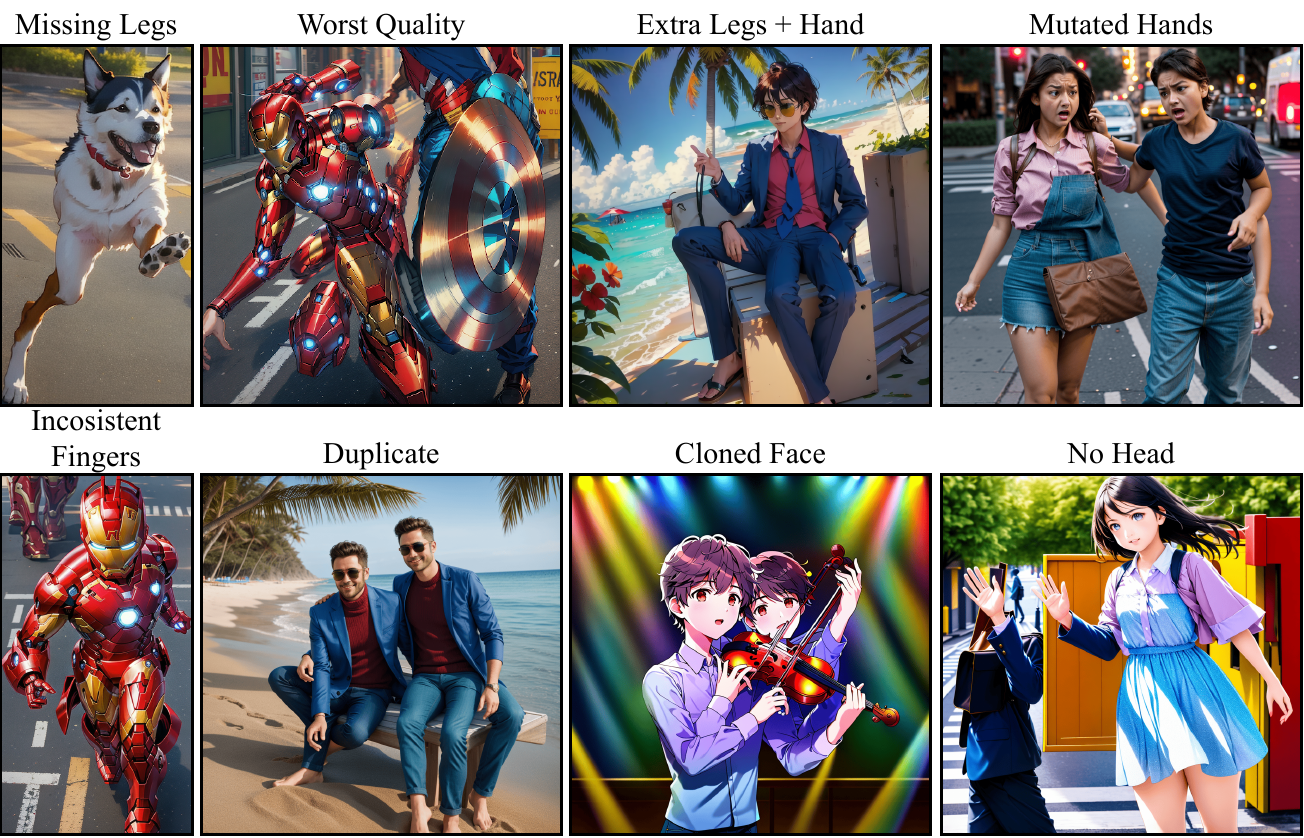}
\vspace{-15pt}
\caption{\textbf{Types of artifacts:} Out of 26 artifacts, these 8 are the most frequent that occur during comic generation.}
\label{fig:arti} 
\end{figure}

Our dataset for VLM fine-tuning consists of 500 images annotated with category and rank (an example of the top 8 categories has been obtained in \cref{fig:arti}). We use LLaVA \cite{liu2024improved} utilized for artifact classification, which contains
a vision encoder, a pre-trained large language model, and a vision-language projector. Here, we perform instructions facilitating fine-tuning as represented in \cite{cao2024synartifact}. For a given synthesized image and question, the vision encoder extracts image features, which are then transformed into the word embedding space by the vision-language projector. These transformed image features are concatenated with language tokens and input into an LLM to generate answers. Cross-entropy loss is computed between the generated answers and the ground truth, allowing the parameters of the vision-language projector and LLM to be updated for both final classification and rank prediction.

\section{Comparison with baseline models}
\label{sec:baseline}
\begin{figure}[!htbp]
\centering
  \includegraphics[width=\linewidth]{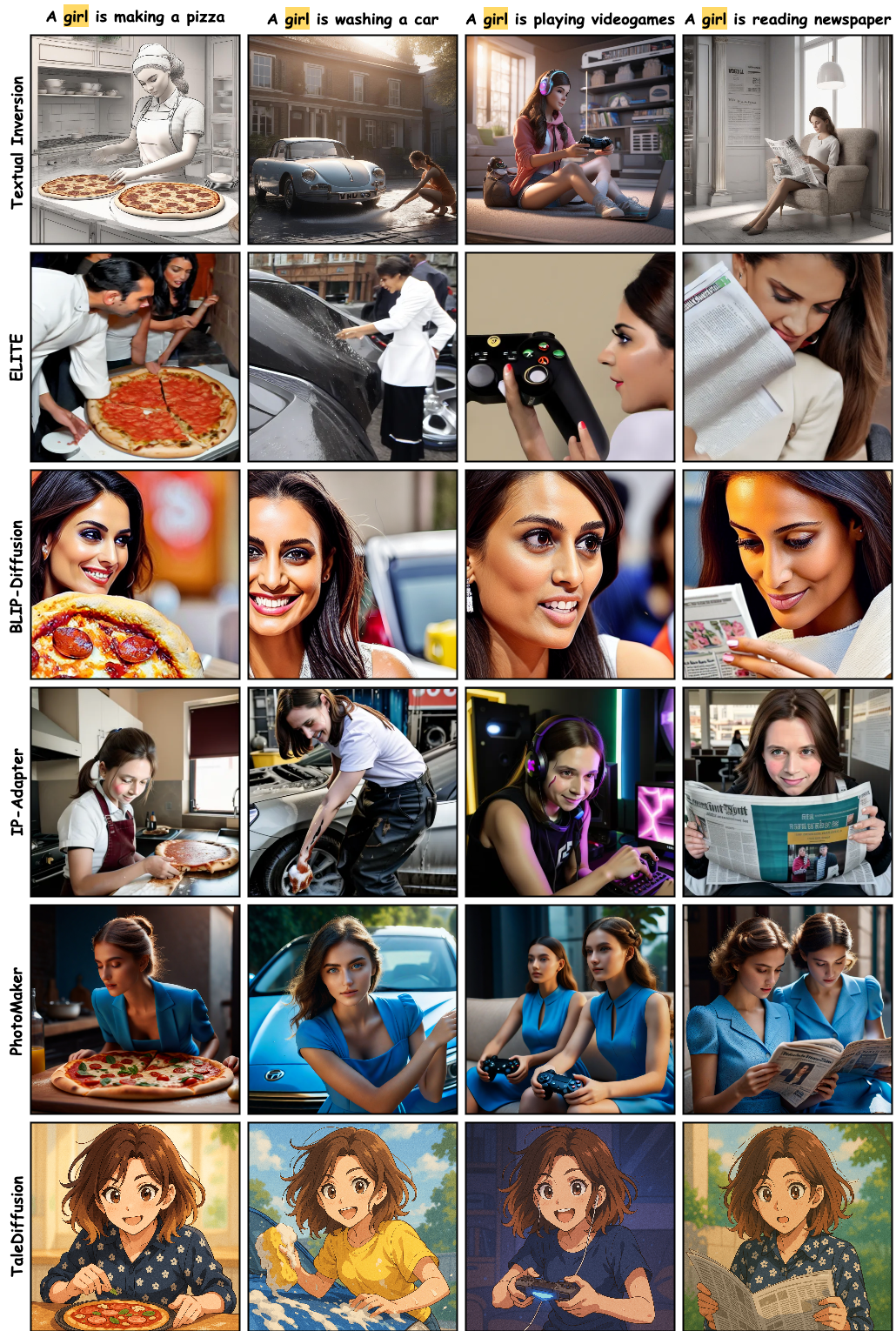}
\caption{Consistent image generation with baseline models for a single character.}
\label{fig:style1} 
\end{figure}

\begin{figure}[!htbp]
\centering
  \includegraphics[width=\linewidth]{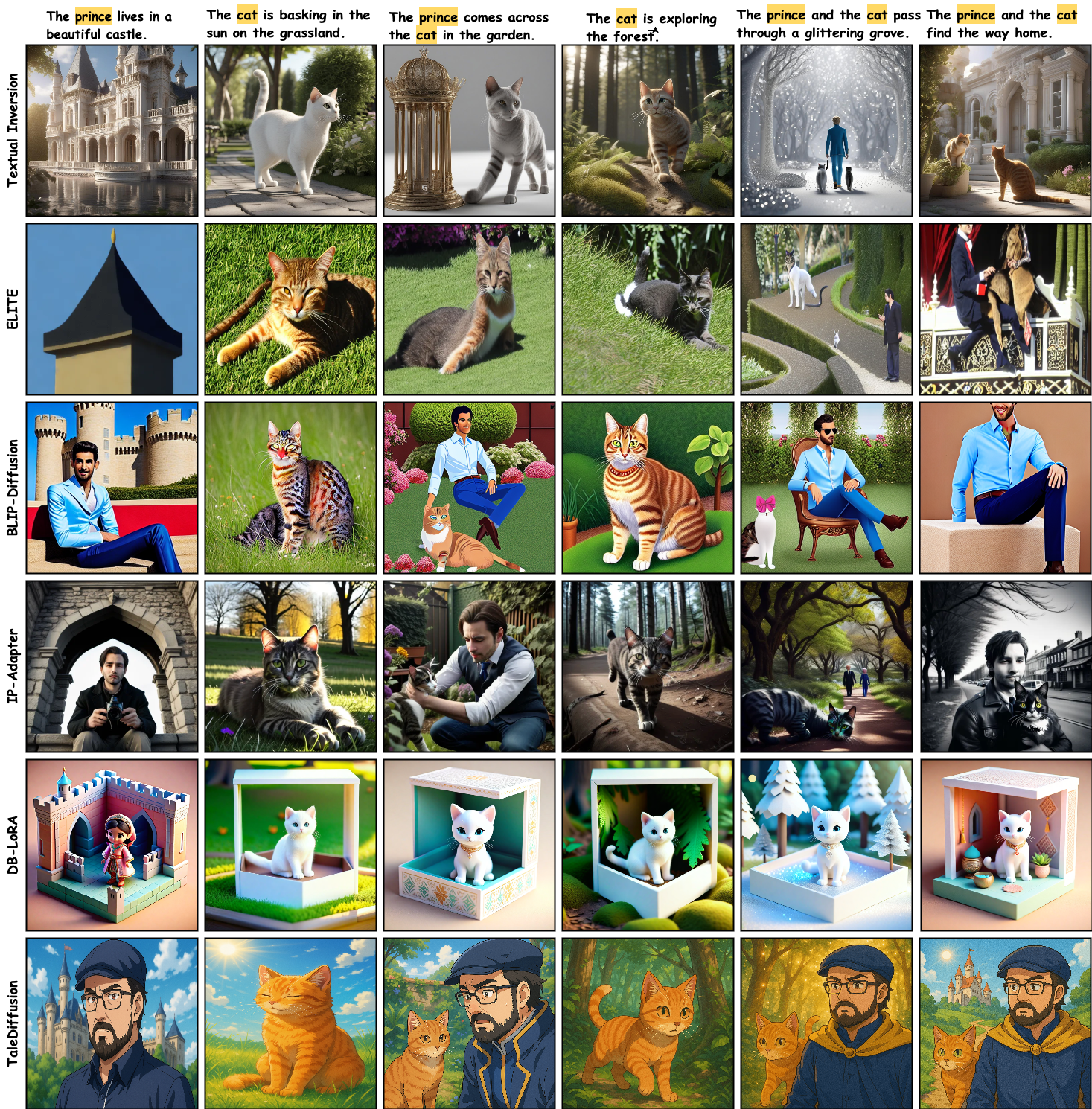}
\caption{Consistent image generation with baseline models for one human and one animal character.}
\label{fig:style2} 
\end{figure}

In this section, we compared TaleDiffusion with the existing state-of-the-art baselines namely Textual Inversion \cite{gal2022image}, ELITE \cite{wei2023elite}, BLIP-Diffusion \cite{li2024blip}, IP-Adapter \cite{ye2023ip}, PhotoMaker \cite{li2024photomaker}, and DB-LoRA \cite{ruiz2023dreambooth} both qualitatively (\cref{fig:style1,fig:style2}) and quantitatively (\cref{tab:001}) for single as well as multiple characters. It has been observed that all baselines method generates artifacts during image synthesis, which proves the effectiveness of mask-based artifacts removal techniques proposed by TaleDiffusion. As depicted in \cref{fig:style1}, Textual Inversion \cite{gal2022image} tries to follow the text prompt during single-character generation, unable to maintain the character consistency. On the other hand, the generated images via ELITE \cite{wei2023elite} pose low quality and mostly suffer from JPEG artifacts. Although BLIP-Diffusion \cite{li2024blip} generates more consistent characters and fewer artifacts compared to the rest, it mostly focuses on the faces. It is unable to generate a proper pose of the characters. Similarly, IP-Adapter \cite{ye2023ip} and PhotoMaker \cite{li2024photomaker} generate consistent images with exact pose it also generates most amount of artifacts (we choose IP-Adapter \cite{ye2023ip} as our baseline, because PhotoMaker \cite{li2024photomaker} is a reference image based generation technique and doesn't work for multiple character generation). In contrast, TaleDiffusion generates consistent, artifact-free characters across stories thanks to the proposed mask guidance, which helps to prevent artifact generation during the denoising process.

\begin{figure}[!htbp]
\centering
  \includegraphics[width=\linewidth]{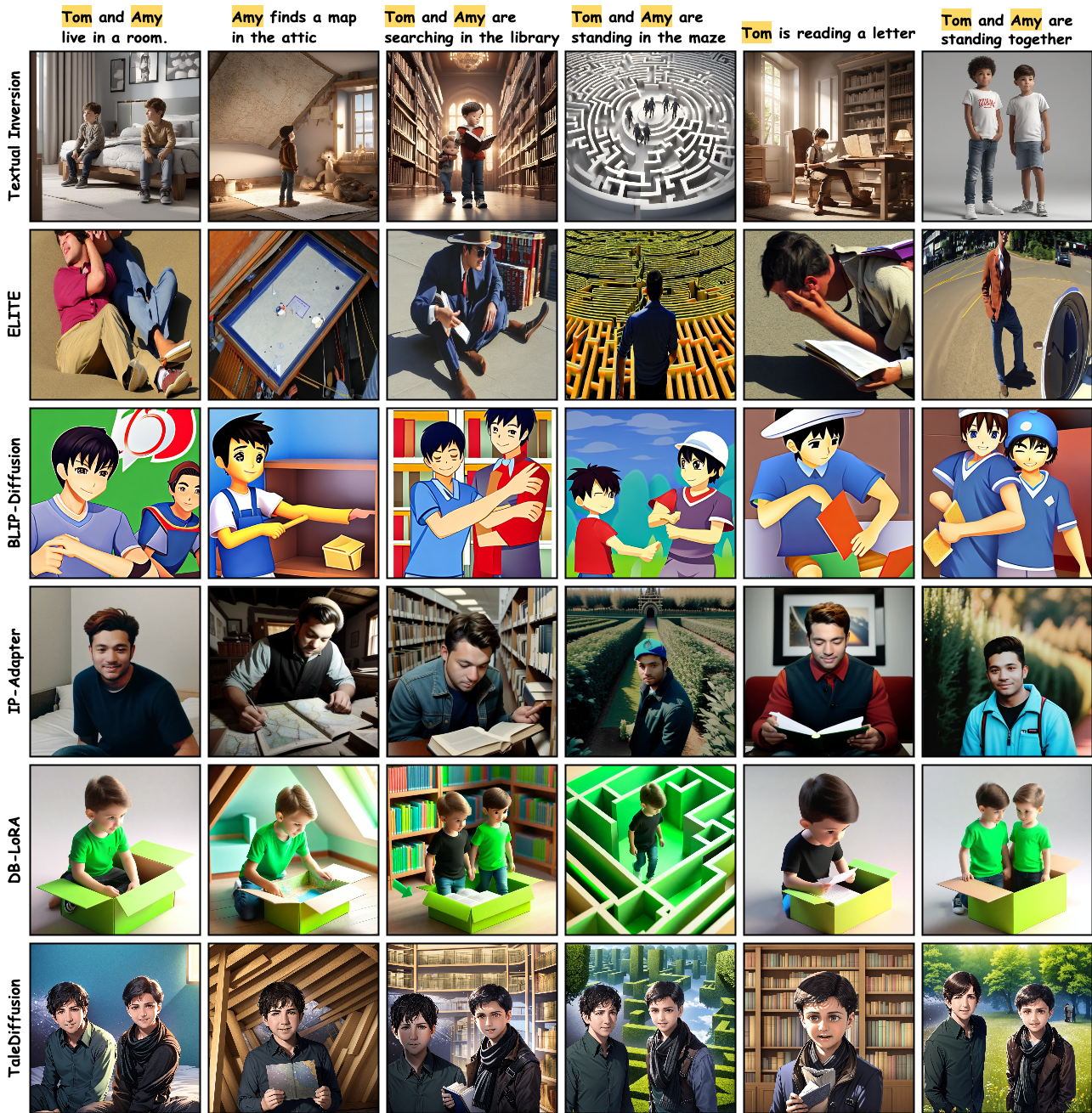}
\caption{Consistent image generation with baseline models for two human characters.}
\label{fig:style3} 
\end{figure}

Similarly, in \cref{fig:style2,fig:style3} Textual Inversion \cite{gal2022image} neither follows the text prompt nor the object count. Elite \cite{wei2023elite} started generating images without any characters. Although BLIP-Diffusion \cite{li2024blip} follows the text prompt unable to maintain consistency. Also, IP-Adapter \cite{ye2023ip} and DB-LoRA \cite{ruiz2023dreambooth} mostly generate panels with single objects. In contrast, TaleDiffusion generates consistent multiple characters with the ICSA-RACA combination. ICSA ensures character consistency in latent guidance $Z_g$ and background $Z_{bg}$ through IP-Adapter \cite{ye2023ip}, while controlling position and size, maintaining consistency for multiple characters. Here, \cref{tab:001} also verifies this conclusion that our TaleDiffusion, although seeing minor performance drops from single to multiple characters due to increased internal noise, achieved high aPSNR and aHPS scores, demonstrating improved diversity and perceptual quality, and outperformed existing approaches across all metrics by a substantial margin.

\begin{table*}[!htbp]
\centering
\caption{Quantitative evaluation on baseline story synthesis techniques}
\label{tab:001}
\resizebox{\textwidth}{!}{
\begin{tabular}{@{}cccccccccc@{}}
\toprule
 & Method/Metrics & Artifacts Score $\downarrow$ & aCCS \cite{zhou2024storydiffusion} $\uparrow$& VQAScore \cite{lin2025evaluating} $\uparrow$ & aBLIP-T \cite{li2022blip} $\uparrow$& aFID \cite{heusel2017gans} $\downarrow$& aPSNR \cite{hore2010image} $\uparrow$& aAes. \cite{Aes2022}$\uparrow$& aHPS \cite{wu2023human}$\uparrow$ \\ \midrule
\multirow{6}{*}{\rotatebox[origin=c]{90}{\textbf{Single char.}}}    & Textual Inversion \cite{gal2022image}   & 0.5723                &  0.2463    &    0.4523     &     0. 3967   &   87.31   &   8.67    &  4.4412     &   0.2318   \\
 & ELITE \cite{wei2023elite}     &    0.8129             &  0.1732    &    0.2132     &    0.2005   &   118.21   &   5.27    &   2.1253    &   0.1726   \\
 & BLIP-Diffusion \cite{li2024blip}      &     0.2112            &   0.6589   &  0.3612       &  0.3112       &     27.43    &   12.19     &   5.9218    &  0.2982    \\
 & IP-Adapter \cite{ye2023ip} &   0.7234              &  0.7168    &   0.7128      & 0.6714        &     38.14   &   11.45    &  4.3452     &   0.2241   \\
 & PhotoMaker \cite{li2024photomaker} &   0.6243              &  0.7911    &   0.7213      & 0.7441        &     24.48   &   13.45    &  6.5432     &   0.2681   \\
 & TaleDiffusion (Ours)      &      \textbf{0.1021}           &    \textbf{0.8088}  &   \textbf{0.8322}      &    \textbf{0.7615}     &    \textbf{11.32}   & \textbf{16.07}       &    \textbf{6.7832}   &  \textbf{0.3276}    \\ \midrule
\multirow{6}{*}{\rotatebox[origin=c]{90}{\textbf{Multi char.}}} & Textual Inversion \cite{gal2022image}    &  0.3245               &  0.2718    &       0.3769         &   0.3754  &  89.76    &  7.28     &   4.4212    &    0.2113  \\
 & ELITE \cite{wei2023elite}       &   0.8927             &  0.1213    &    0.1322     &    0.1005   &   123.12   &   4.72    &   2.1135    &   0.1562    \\
 & BLIP-Diffusion \cite{li2024blip} &   0.2791            &   0.6098   &  0.2937       &  0.3001       &     59.34    &   10.27     &   4.2981    &  0.2128   \\
 & IP-Adapter \cite{ye2023ip} &   0.7013              &  0.6542    &   0.6182      & 0.5914        &     41.24   &   10.54    &  4.5243     &   0.2114   \\
 & DB-LoRA \cite{ruiz2023dreambooth} &   0.1324              &  0.6912    &   0.6132      & 0.594        &     22.18   &   14.11    &  5.4523     &   0.2318   \\
 & TaleDiffusion (Ours)      &      \textbf{0.1103}           &     \textbf{0.7992}        &    \textbf{0.7864}     &  \textbf{0.7622}   &   \textbf{17.53}   &  \textbf{16.76}     &   \textbf{6.5412}    &    \textbf{0.3301}  \\ \bottomrule 
\end{tabular}
}
\end{table*}

\section{Story with different style}
\label{sec:style}
\begin{figure*}[!htbp]
\centering
  \includegraphics[width=\textwidth]{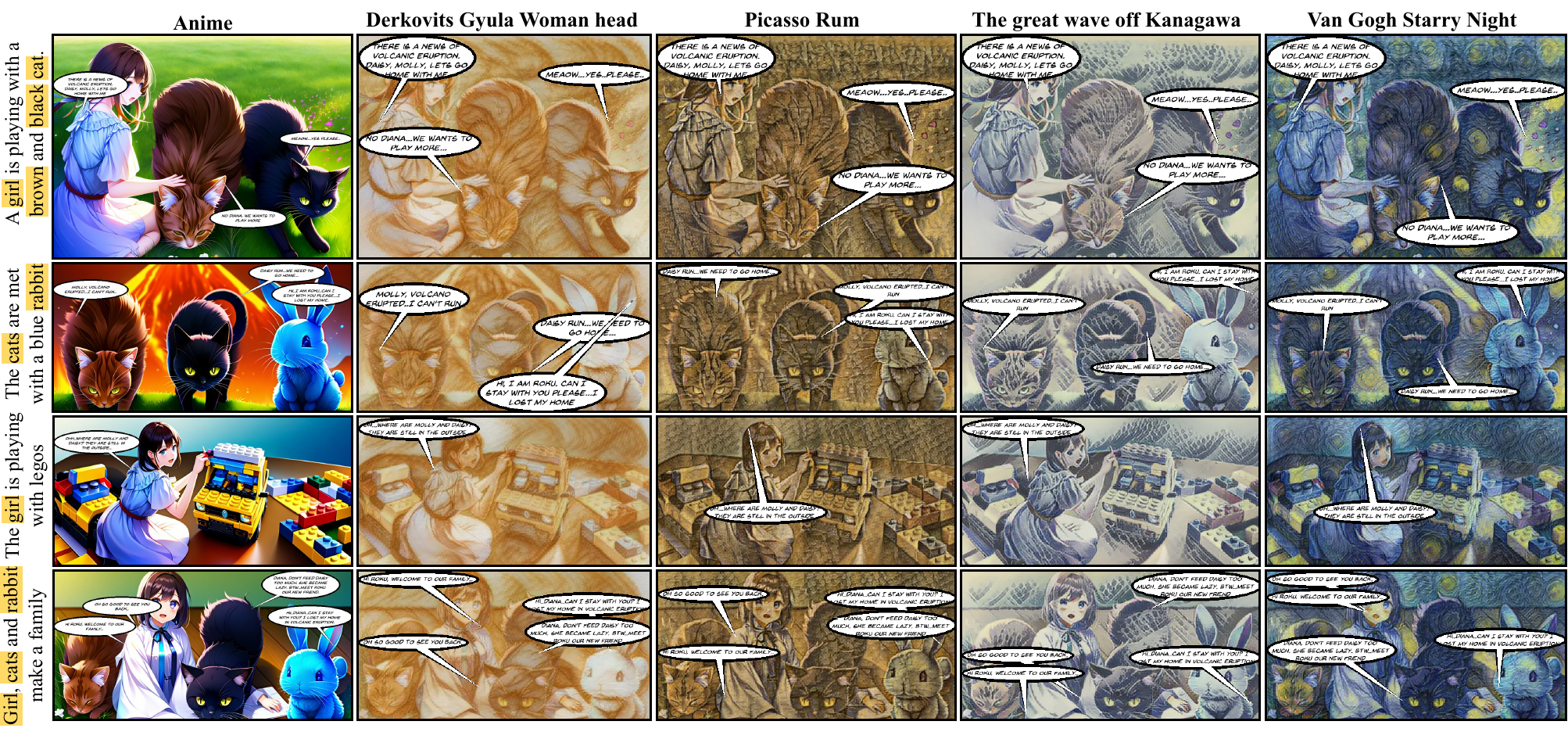}
\caption{\textbf{Robustness of dialogue rendering against different styles:} It has been observed that the panels generated by TaleDiffusion are not only consistent across different styles also the bubbles are correctly assigned to the characters against the noise caused by styles.}
\label{fig:style} 
\end{figure*}

This experiment assesses the robustness of TaleDiffusion across different styles and evaluates the noise tolerance of the proposed dialogue rendering technique. We applied style transfer, based on paintings by renowned artists (\eg, Van Gogh, Derkovits Gyula), following the approach in \cite{gatys2017controlling}. As shown in \cref{fig:style}, dialogue bubbles remain correctly assigned to each character, even though bubble positions slightly shift due to style-induced noise, without covering character faces or affecting panel quality.

Also, the character consistency and the concept of the story are maintained across different styles (aCCS: 0.7401, aCLIP-T: 0.7312). 
We have performed these experiments for 10 stories with multiple characters ($\geq$ 2) each containing four panels across fourteen different styles and getting an average R@(\#text) of 0.6125. This is because CLIPSeg \cite{luddecke2022image} assigned the rendered text to the characters by considering its physical attributes and action, along with its color, which makes our model robust against the noise caused by the styles.

\begin{figure*}[!htbp]
\centering
\includegraphics[width=\textwidth]{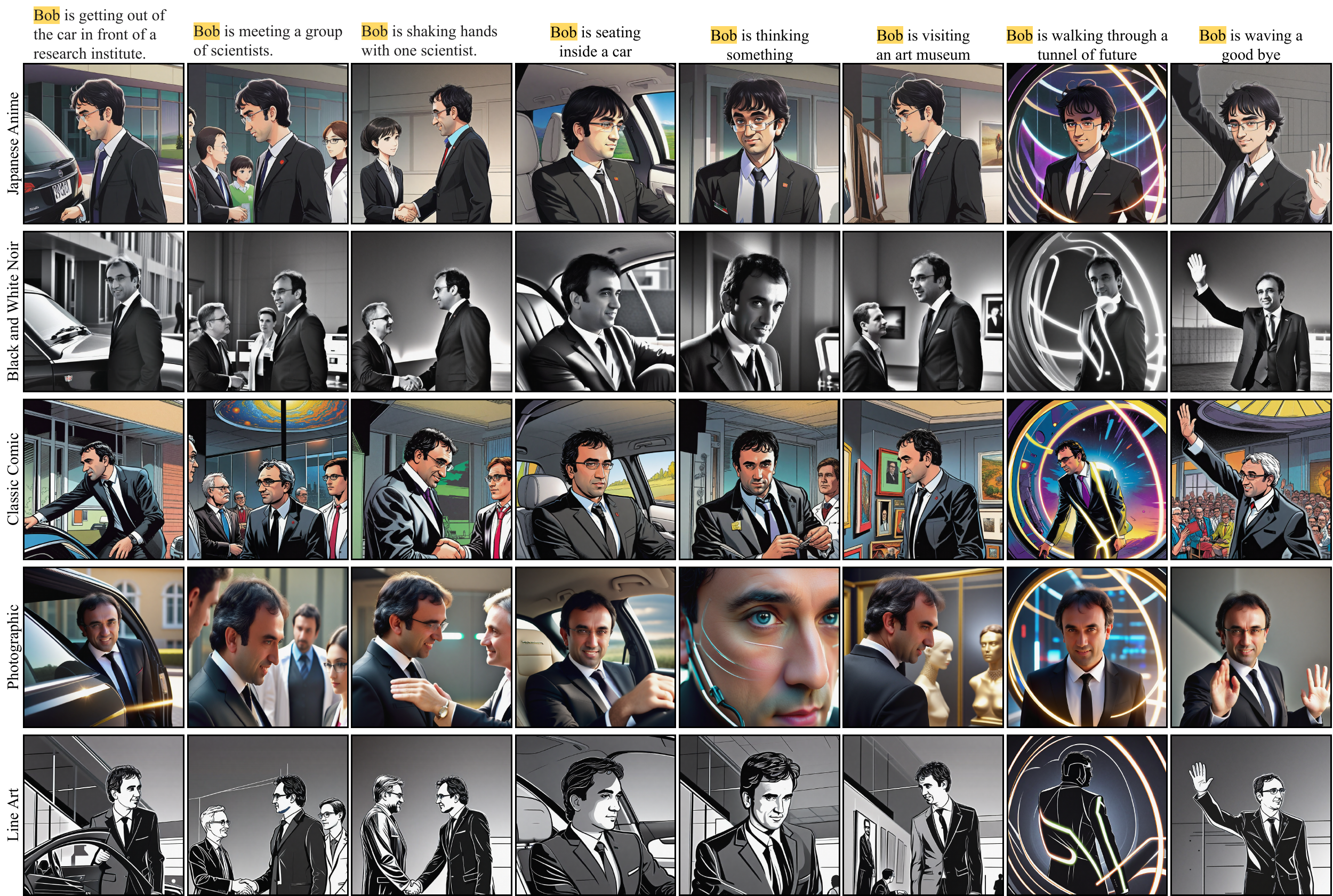}
\caption{\textbf{Stylization:} We can generate the same story with different styles, maintaining character consistency and frame coherency.}
\label{fig:exp_on_style2} 
\end{figure*}

Similarly, we have tried to change the add different style on the diffusion U-Net \footnote[3]{https://civitai.com/} in order to generate the same story with user given style preserving content and character consistency. The results are reported in \cref{fig:exp_on_style2}. It has been observed that TaleDiffusion is able to maintain character consistency across different styles.

\section{More details on Human Evaluation}
\label{sec:human}
\begin{figure}[!htbp]
\centering
\includegraphics[width=\columnwidth]{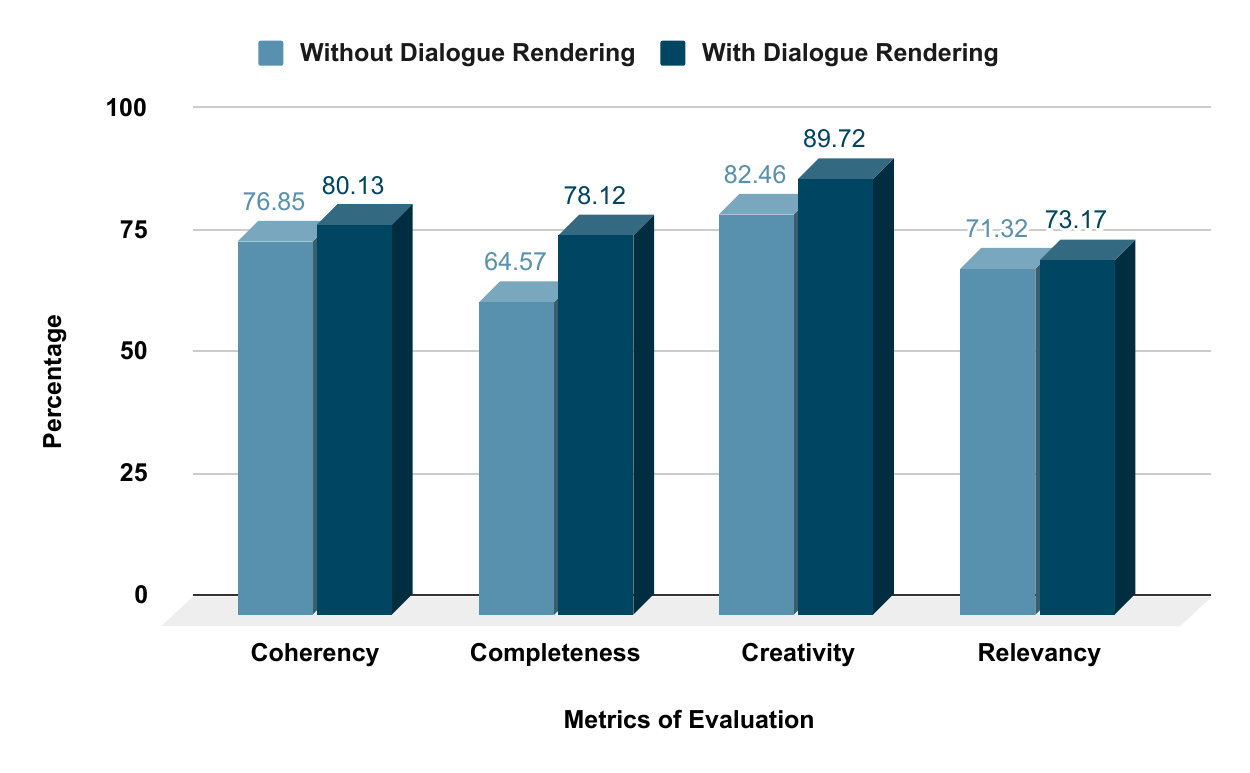}
\caption{\textbf{Effectiveness of Dialogue Rendering:} It has been observed that dialogues improved the story generation over the images in all perspectives.}
\label{fig:chart} 
\end{figure}

This section reports the user study we have conducted in order to evaluate the credibility and widespread adoption of synthesized comics. Here we ask 22 people to evaluate our work in the following settings: in the first setting, we provide the stories generated by TaleDiffusion with and without the dialogue rendering and ask them to judge it on 10 generated comics based on creativity, completeness, coherency, and relevancy. The result has been reported in \cref{fig:chart}. It has been observed that dialogues make the story complete and relevant as dialogues add depth and context to a story by revealing characters's thoughts, emotions, and relationships, which images alone cannot fully convey. It also enhances coherence and creativity by driving the narrative forward and engaging readers through meaningful interactions.

\begin{figure}[!htbp]
\centering
\includegraphics[width=\linewidth]{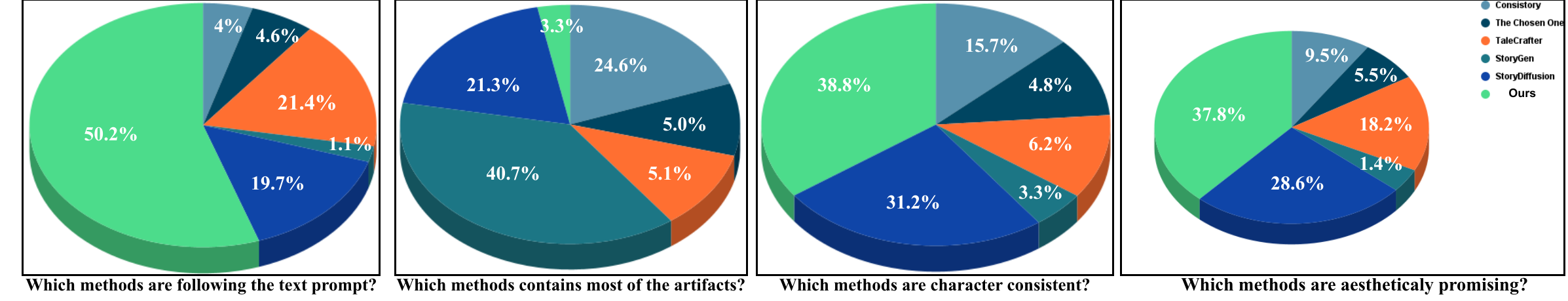}
\caption{\textbf{Story Generation:} TaleDiffusion gets the maximum vote from the user in all the categories, whereas StoryGen \cite{liu2024intelligent} gets the least. on the other hand, StoryDiffusion \cite{zhou2024storydiffusion} give some competition to TaleDiffusion in character consistency.}
\label{fig:user} 
\end{figure}

In the second experiment, we generated 5 comics (2 single and 3 multiple-character comics) with TaleDiffusion as well as the existing state-of-the-art approaches namely Consistory \cite{tewel2024training}, The Chosen One \cite{avrahami2024chosen}, TaleCrafter \cite{talecrafter2024}, StoryGen \cite{liu2024intelligent}, Story Diffusion \cite{zhou2024storydiffusion} and ask the user following four questions: 1) Which method is correctly following the text prompt? 2) Which method contains most of the artifacts? 3) Which method is aesthetically more pleasing? and 4) Which method is more character consistent? The user can vote for at most two methods, and the result has been reported in \cref{fig:user}.

From \cref{fig:user}, it has been observed that the panels generated by StoryGen \cite{liu2024intelligent} contain most of the artifacts. Also, it neither follows the text prompt nor maintains character consistency and the generated images are not aesthetically very promising. StoryDiffusion \cite{zhou2024storydiffusion} stood the second position in all of these above-mentioned categories except artifact generation and it generates artifacts in 21.3\% panels. In contrast, all the users agreed that the comics synthesized by CraftSVG properly followed the complex text prompt, generating fewer artifacts, consistent, and aesthetically promising which consolidated our claim in this paper.

\section{Limitations and Future Works}
\label{sec:fail}
\myparagraph{Limitations:} 
The proposed TaleDiffusion mostly suffered from the following three limitations: Firstly, the inference time. As we are using low-rank adaptation for multi-character customization it takes $\approx$ 35 seconds per character customization. We have compared TaleDiffusion with state-of-the-art approaches for single-frame generation with a single character in \cref{tab:fail}. As none of the methods fail to generate multiple characters and the inference time of a single panel with TaleDiffusione depends on how many characters we are trying to generate in a single panel.

\begin{table}[!htbp]
\centering
\caption{Inference time comparison with the existing methods}
\label{tab:fail}
\resizebox{\columnwidth}{!}{
\begin{tabular}{@{}cccccc@{}}
\toprule
Methods         & DB-LoRA \cite{ruiz2023dreambooth}   & Textual Inversion \cite{gal2022image} & IP-Adapter \cite{ye2023ip} & BLIP-Diffusion \cite{li2024blip} & TaleDiffusion \\ \midrule
Inf. Time (sec) & 312        & 455               & 12.1       & 9.4            & 34.7      \\ \midrule \\ \midrule
Methods         & Consistory \cite{tewel2024training} & The Chosen One \cite{avrahami2024chosen}   & StoryGen \cite{liu2024intelligent}  & StoryDiffusion \cite{zhou2024storydiffusion} & TaleDiffusion \\ \midrule
Inf. Time (sec) & 15.6       & 524.2             & 16.8       & 6.7            & 34.7      \\ \bottomrule
\end{tabular}
}
\end{table}

As reported in \cref{tab:fail}, the Chosen One \cite{avrahami2024chosen} has the highest inference time as it involves image clusterization. On the other hand, StoryDiffusion \cite{zhou2024storydiffusion} is the fastest method as it doesn't involve any training or fine-tuning. Similarly, IP-Adapter \cite{ye2023ip}, and BLIP-Diffusion \cite{li2024blip} have very low inference time as they are also training-free methods. On the other hand, compared to DB-LoRA \cite{ruiz2023dreambooth}, which takes 455 seconds for a single character inference our methods optimize it with gradient fusion and provide better results within $\approx$ 35 seconds (Note: all the experiments have been performed using a single NVIDIA A100 GPU). Secondly, TaleDiffusion is unable to generate more than four characters in a single frame due to the applied constraints in layout generation (every bounding box shouldn't have a lesser area than 1/4th of the total canvas area, otherwise it is hard to generate a prominent character face). Lastly, sometimes the character's clothes will be inconsistent, which can be solved by trying multiple seeds. Otherwise, we need to fine-tune the model with cross-domain consistency loss \cite{wang2023cdac} in order to maintain consistency in clothing level.

\myparagraph{Broader Impacts:} It is crucial to remain mindful of potential risks associated with these models, including disseminating misinformation, potential for abuse, and introducing biases. Broader impacts and ethical considerations should be thoroughly addressed and studied to responsibly harness such models' capabilities.

\begin{figure}[!htbp]
\centering
\includegraphics[width=\columnwidth]{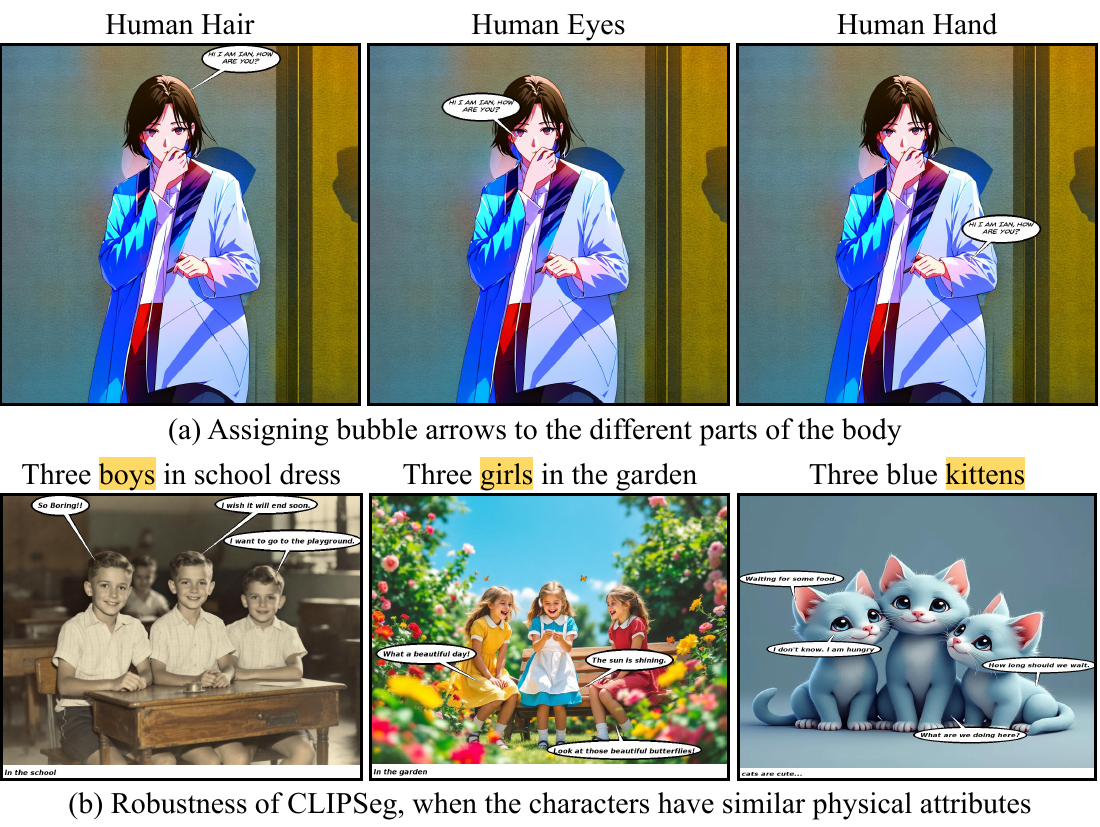}
\caption{\textbf{Ablation of CLIPSeg:} We can provide different text prompts mentioning human body parts (i.e., eyes, hair, hands, and so on) to assign the bubble to the character even though the characters have similar physical attributes.}
\label{fig:ab_clip} 
\end{figure}

\begin{figure*}[!htbp]
\centering
\includegraphics[width=\textwidth]{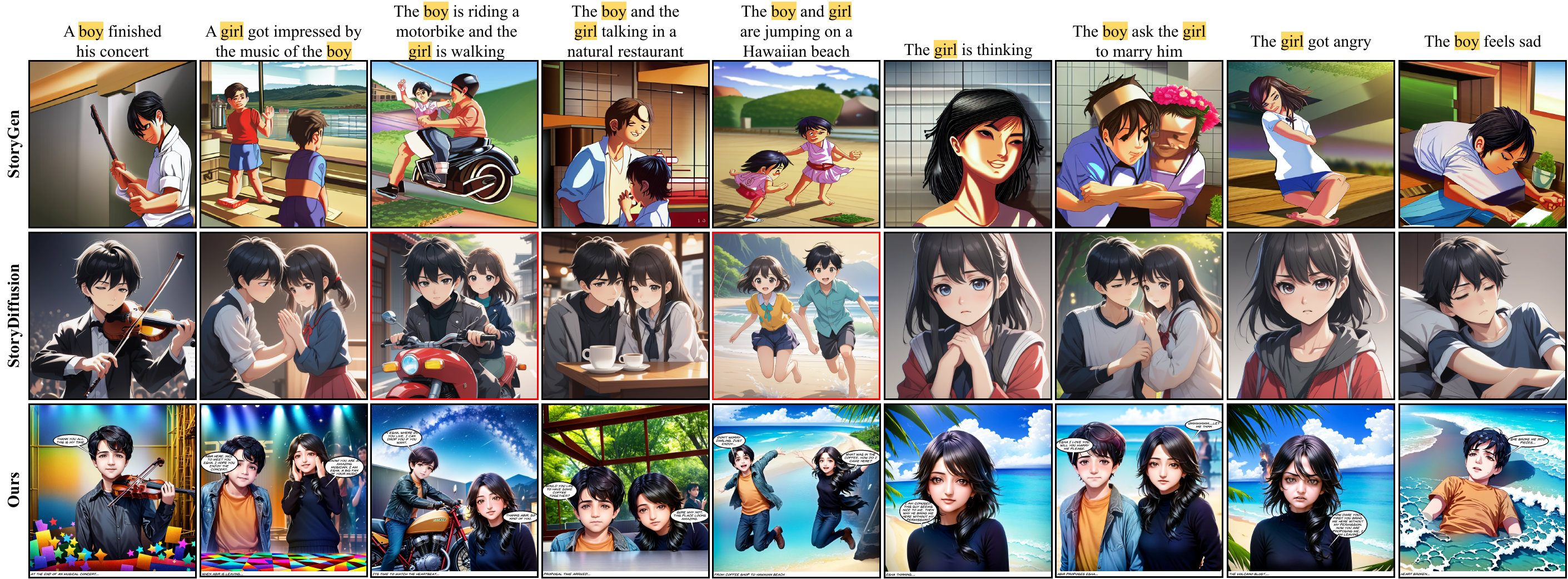}
\caption{\textbf{Qualitative Evaluation:} StoryGen neither has consistency nor follows the text prompt. Although StoryDiffusion follows the text prompt, it fails to maintain character as well as background consistency and generates artifacts. In contrast, TaleDiffusion follows the text prompt and maintains character as well as background consistency.}
\label{fig:supp_comp} 
\end{figure*}

\begin{figure*}[!htbp]
\centering
\includegraphics[width=\textwidth]{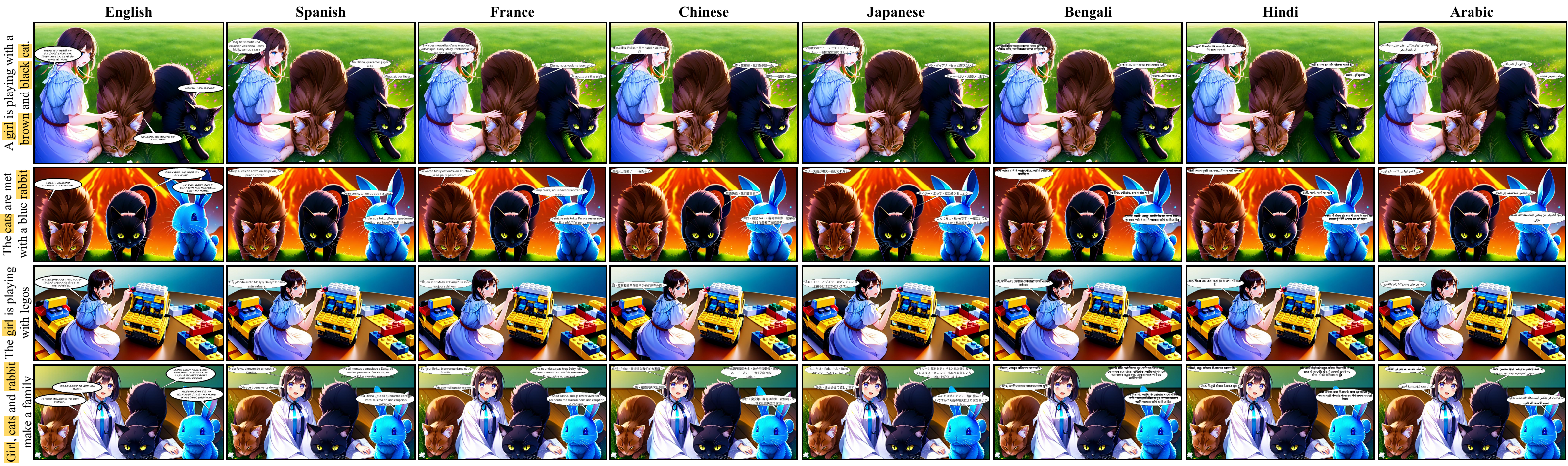}
\caption{\textbf{Experiments on Multilingualism:} We can generate the same story in different languages, maintaining character consistency and frame coherence.}
\label{fig:exp_on_style} 
\end{figure*}

\section{Qualitative Examples}
\label{sec:lecun}
\begin{figure*}[t]
\centering
\includegraphics[width=\textwidth]{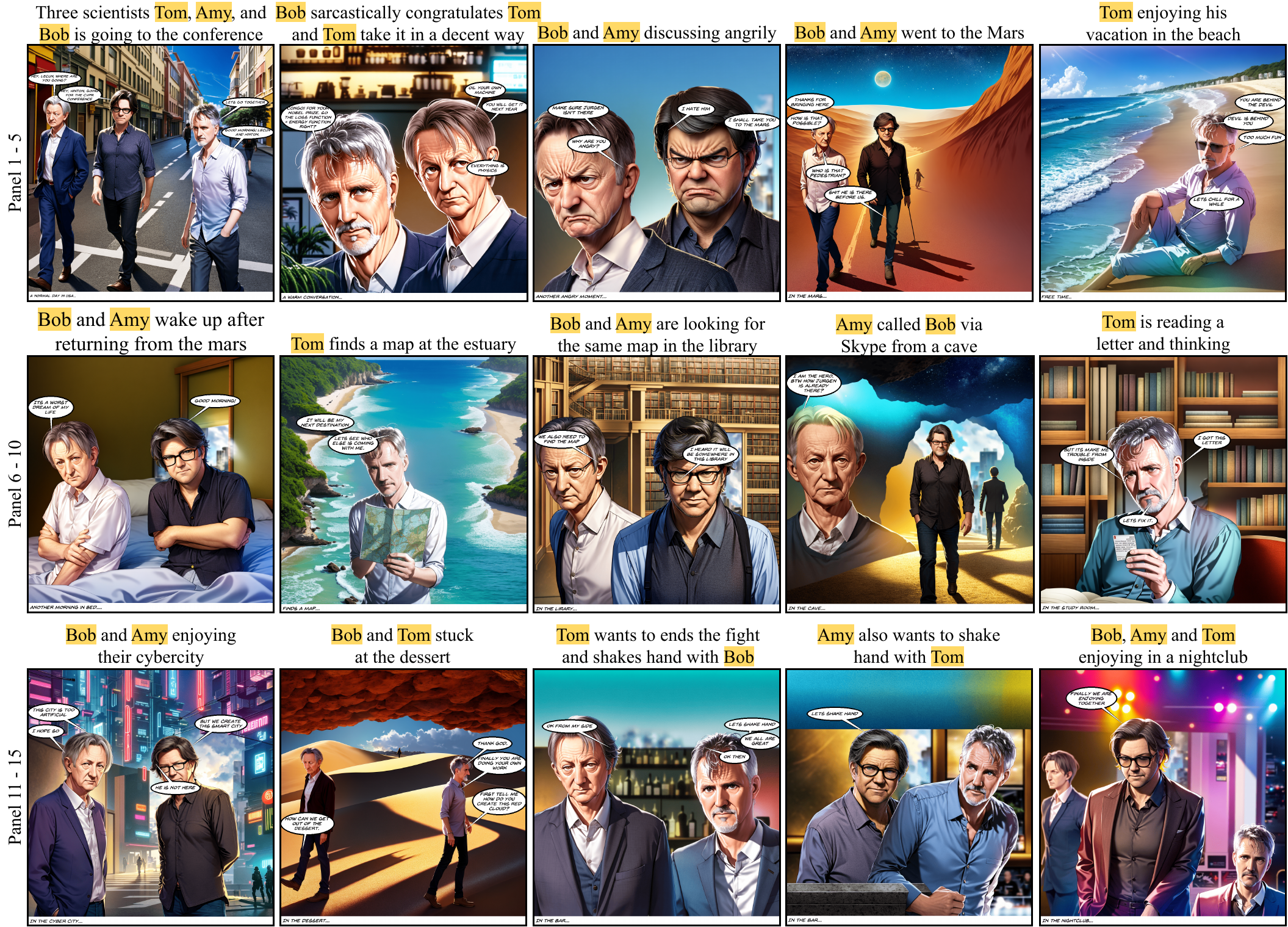}
\caption{\textbf{A journey from hatred to friendship:} Three scientists hate each other, met at a conference. They compete against each other for new inventions. Ultimately they become friends and start enjoying life. Please zoom in for better visualization.}
\label{fig:ex1} 
\end{figure*}

\begin{figure*}[!htbp]
\centering
\includegraphics[width=\textwidth]{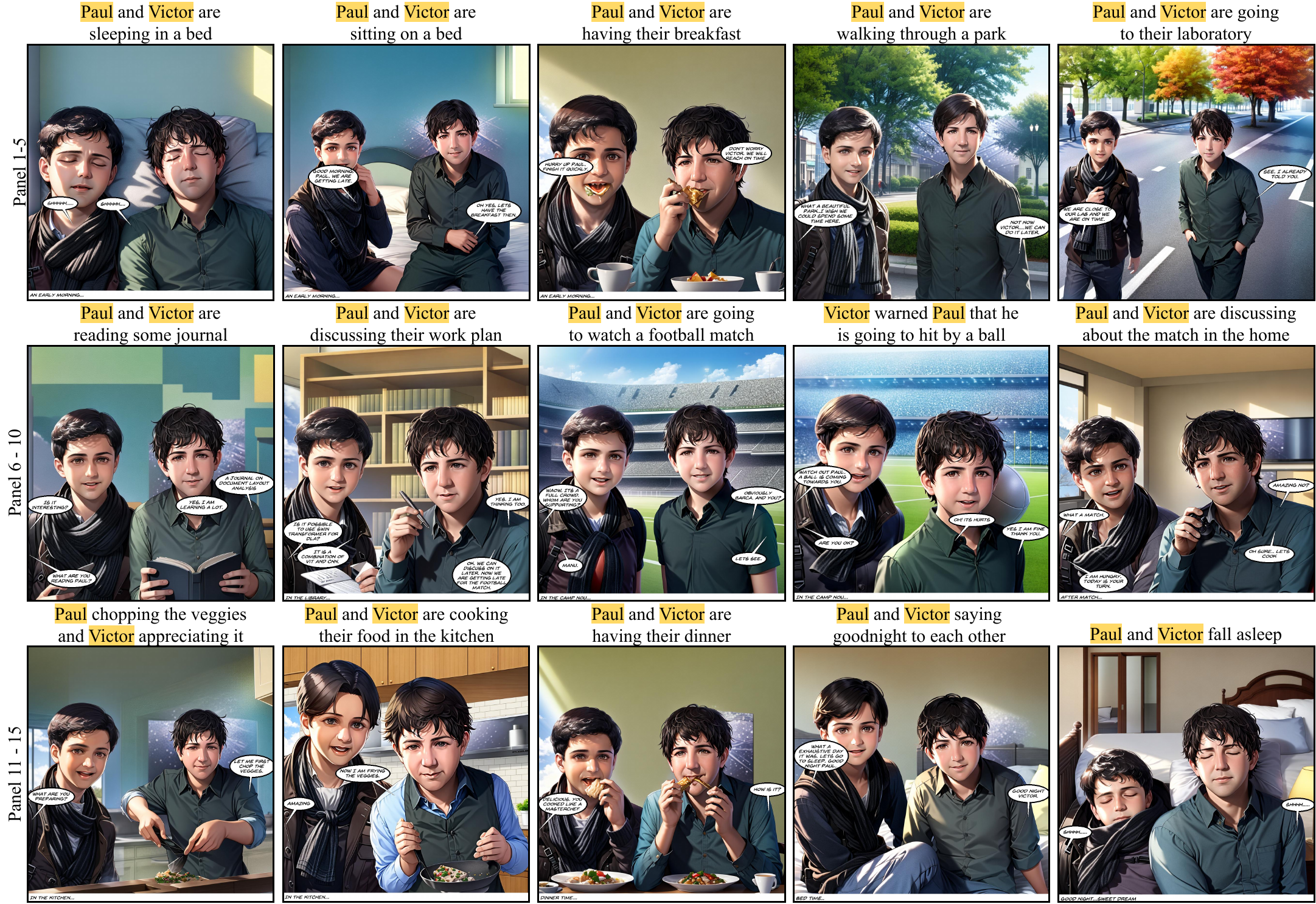}

\caption{\textbf{A daily life of two friends:} Paul and Victor live together. Every day they wake up, do their work, enjoy life, and go to sleep again. Please zoom in for better visualization.}
\label{fig:ex3} 
\end{figure*}

\begin{figure*}[!htbp]
\centering
\includegraphics[width=\textwidth]{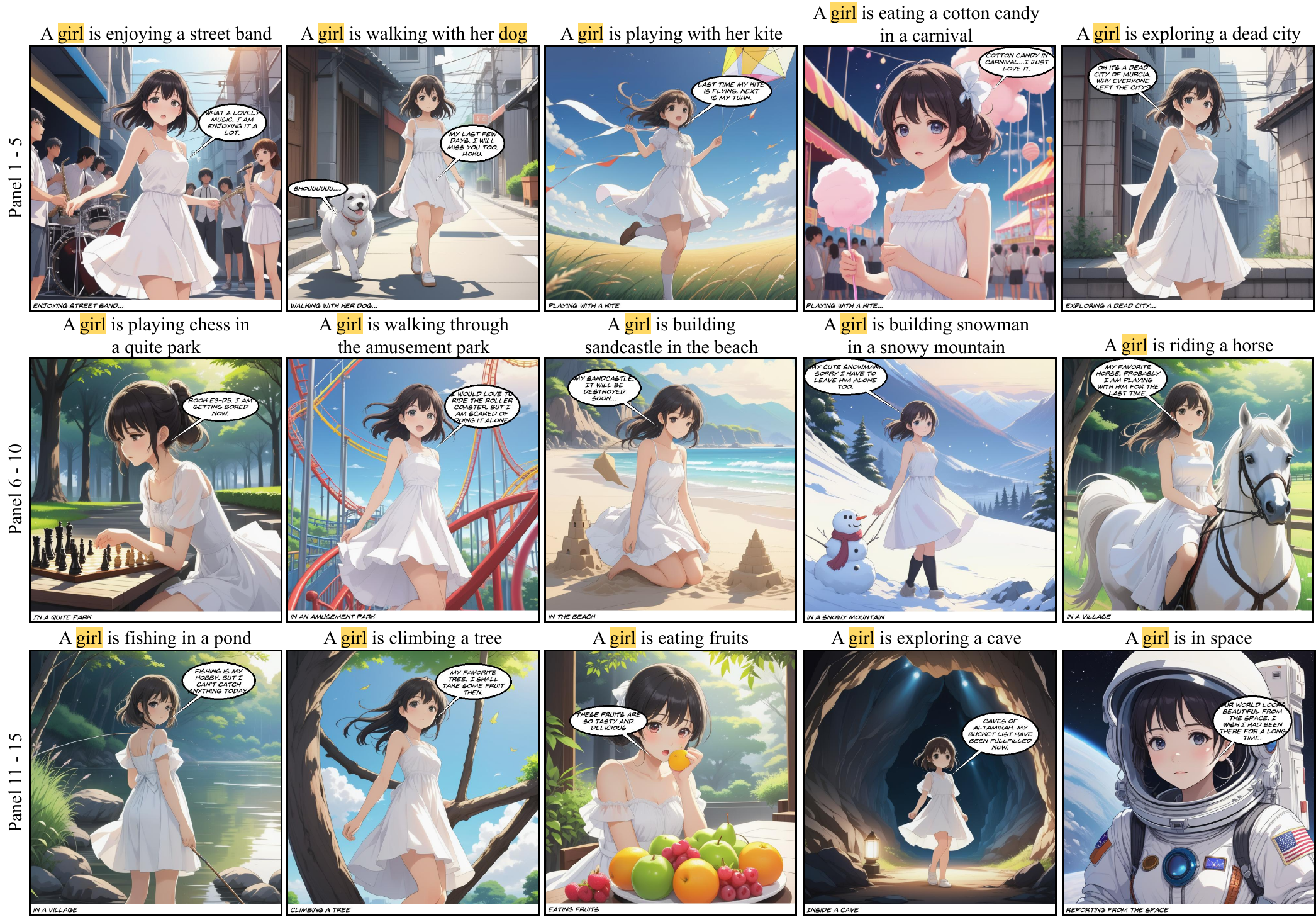}

\caption{\textbf{The last few days of a girl} A girl is about to go to space. Before that, she is enjoying her life at its best. Please zoom in for better visualization.}
\label{fig:ex4} 
\end{figure*}

\begin{figure*}[!htbp]
\centering
\includegraphics[width=\textwidth]{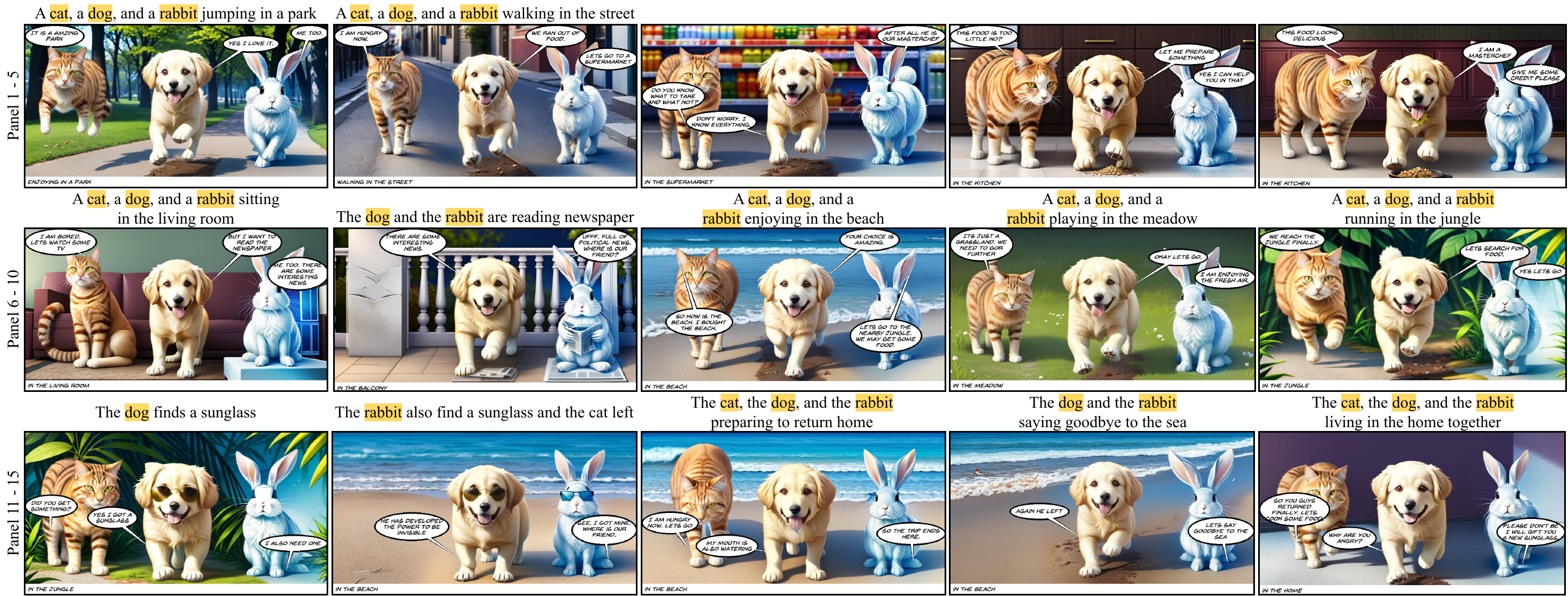}
\caption{\textbf{Sharing is Caring:} A cat, a dog, and a rabbit live together, play together, eat, travel, and are always happy together. Please zoom in for better visualization.}
\label{fig:ex5} 
\end{figure*}

In this section, we have obtained some full-length comics (15 panels) generated by TaleDiffusion with varying numbers of characters and their corresponding activities. As shown in \cref{fig:ex1}, TaleDiffusion tries to create some essence of "video call through Skype" by creating an abstract character (row 2, column 4 of \cref{fig:ex1}). Similarly, it creates some zoom-in effects on the face whenever necessary. Similarly, in the second story (see \cref{fig:ex3}), "A Daily Life of Two Friends" TaleDiffusion correctly depicts activities like sleeping, thinking, and so on. On the other hand, \cref{fig:ex4} is a single-character story, where TaleDiffusion perfectly aligned the main character (a girl in a white dress) with auxiliary characters (dog, horse, street band), and so on. Also, in the last panel (row 3, column 5 of \cref{fig:ex4}) it perfectly fits the astronaut suit on the character. The last story is about a cat, a dog, and a rabbit (see \cref{fig:ex5},) although there are not many pose variations (only standing, running, jumping, and so on) it perfectly creates the effect of reading books and wearing sunglasses. Besides that, all the panels generated in \cref{fig:ex1,fig:ex3,fig:ex4,fig:ex5} are not only artifact-free and consistent but also the dialogue bubbles are correctly assigned to the characters, which improves readability, coherence, and the completeness of the overall story.

\section{Ablation of CLIPSeg}
\label{sec:code}
CLIPSeg, as proposed in \cite{luddecke2022image}, is a powerful segmentation model that accepts a text prompt and an image as inputs, enabling it to identify and segment the image region corresponding to the given textual description. This capability has been effectively utilized for bubble assignment in our approach, where we input specific human body parts—such as "hair," "eyes," "lips," or "hand"—as text prompts. CLIPSeg then assigns the speech or thought bubble to the segmented region of the image matching the prompt. The effect of varying the text prompt on bubble assignment is illustrated in \cref{fig:ab_clip}. It is evident that CLIPSeg accurately associates the bubbles with the intended regions when we specify these regions in the text prompt. For example, when "hair," "eyes," or "hand" were used as prompts, the model successfully positioned the bubbles over the respective regions. This behavior highlights the flexibility and robustness of CLIPSeg in understanding and aligning textual input with image content.

Not only that, when the characters have similar facial attributes and dresses, CLIPSeg successfully assigns the bubbles as it also considers the physical description, personality, as well as actions. Also, we can assign multilingual bubbles to the characters by enabling the UTF-8 encoding (see \cref{fig:exp_on_style}).

\myparagraph{Problem Definition}
Given an image $I$ with dimensions $W \times H$, a set of characters $\{c_1, c_2, \dots, c_n\}$, and for each character $c$, a physical description $D_c$ and dialogue text $T_c$, the goal is to determine optimal positions for speech bubbles.

\myparagraph{Definitions}
\begin{itemize}
    \item $\text{CLIPSeg}(I, \text{text})$ returns:
    \begin{itemize}
        \item A head location $(x, y)$.
        \item A heatmap $\Lambda$ normalized so that $\Lambda : I \to [0,1]$.
    \end{itemize}
    \item $\text{wrap}(T, W_{\max})$ wraps text $T$ within a width $W_{\max}$.
    \item $f_{\text{size}}(T, \text{font})$ returns the size $(w_t, h_t)$ of $T$.
\end{itemize}

\begin{algorithm}[!htbp]
    \caption{Bubble Placement System}
    \label{algo2}
    \begin{algorithmic}
        \For{each character $c$ in the frame}
            \State $(x, y, \Lambda) \gets \text{CLIPSeg}(I, \{D_c\} \cup \{\text{''Face''}\})$
            \State $L_0 \gets \Lambda(x, y)$
            \State $d \gets \begin{cases} -1, & \text{if } x < \frac{W}{2} \\ +1, & \text{otherwise} \end{cases}$
            \State $(x_e, y_e) \gets (x, y)$
            \While{$(x_e, y_e)$ inside bounds and $\Lambda(x_e, y_e) \geq L_0 - 0.1 \cdot |L_0|$}
                \State $x_e \gets x_e + 2d, \quad y_e \gets y_e - 0.1$
            \EndWhile
            \State $T_{\text{wrapped}} \gets \text{wrap}(T_c, W_{\max})$
            \State $(w_t, h_t) \gets f_{\text{size}}(T_{\text{wrapped}}, \text{font})$
            \State $W_{\text{bubble}} \gets w_t + p, \quad H_{\text{bubble}} \gets h_t + p$
            \If{$x < \frac{W}{2}$}
                \State $x_b \gets x - 60 - W_{\text{bubble}}$
            \Else
                \State $x_b \gets x + 60$
            \EndIf
            \State $y_b \gets y - 60 - H_{\text{bubble}}$
            \If{$y_b < H_{\text{min}}$}
                \State $y_b \gets 0$
            \EndIf
            \If{$|y_b - y_e| < \epsilon$}
                \State $y_b \gets y_e - H_{\text{bubble}} - 20$
            \EndIf
            \State $x_b \gets \max(0, \min(x_b, W - W_{\text{bubble}}))$
            \State $y_b \gets \max(0, \min(y_b, H - H_{\text{bubble}}))$
            \State $B_c \gets (x_b, y_b, W_{\text{bubble}}, H_{\text{bubble}}, (x_e, y_e), T_{\text{wrapped}}, (x, y))$
        \EndFor
        
        \For{each pair $(B_i, B_j)$ in bubbles}
            \If{$B_i$ covers $B_j$'s head}
                \State Adjust $B_i$: $x_b \gets x_b + 80 \cdot \text{sign}(W/2 - x_b)$
                \State $y_b \gets y_e + H_{\text{bubble}} + 80$
            \EndIf
            \If{$B_i$ and $B_j$ overlap}
                \State $y_b^j \gets B_i \cdot y_b + H_{{\text{bubble}}_i}$
            \EndIf
        \EndFor
        
        \For{each bubble $B$}
            \State Draw ellipse at $(x_b, y_b)$ with dimensions $(W_{\text{bubble}}, H_{\text{bubble}})$
            \State Compute $\theta = \arctan\left(\frac{y_e - c_y}{x_e - c_x}\right)$
            \State Compute arrow points $P_1, P_2$
            \State Draw triangle with vertices $P_1, P_2, (x_e, y_e)$
            \State Render $T_{\text{wrapped}}$ centered in the bubble
        \EndFor
    \end{algorithmic}
\end{algorithm}

\begin{algorithm}
\caption{Get Location with CLIPSeg}
\label{algo3}
\begin{algorithmic}[1]
    \Require Image \( I \), Text Prompts \( T \), Used Locations \( L \) (default: empty)
    \Ensure Estimated coordinates \( (x, y) \), Scaled logits \( S \)
    
    \State \( T \gets T \cup \{\text{"Human Head"}\} \) \Comment{Append "Human Head" to text list}
    \State \( \text{inputs} \gets \text{Processor}(T, I, \text{padding}=\text{True}) \) \Comment{Preprocess inputs}
    \State \( O \gets \text{Model}(\text{inputs}) \) \Comment{Pass through segmentation model}
    
    \State Normalize second output:
    \[
    O_2' \gets \frac{O_2 - \min(O_2)}{\max(O_2) - \min(O_2)}
    \]

    \State Compute logical AND:
    \[
    C \gets O_1 \cdot O_2'
    \]

    \State Find maximum probability location:
    \[
    (y, x) \gets \arg\max(C)
    \]

    \State Scale coordinates:
    \[
    x \gets x \cdot \frac{W_I}{W_C}, \quad y \gets y \cdot \frac{H_I}{H_C}
    \]

    \State Resize logits:
    \[
    S \gets \text{Interpolate}(C, (H_I, W_I))
    \]

    \ForAll { \( (x_L, y_L) \in L \) }
        \State Compute Euclidean distance:
        \[
        d \gets \sqrt{(x_L - x)^2 + (y_L - y)^2}
        \]
        
        \If { \( d < 400 \) }
            \State Define mask \( M \) centered at \( (x, y) \) with size \( 100 \times 100 \)
            \State Reduce logits in \( M \) by \( 50\% \):
            \[
            S[M] \gets S[M] \cdot 0.5
            \]
            \State Recompute new maximum location:
            \[
            (y, x) \gets \arg\max(S)
            \]
        \EndIf
    \EndFor

    \State \Return \( (x, y), S \)
\end{algorithmic}
\end{algorithm}

Out of these cases, the bubble assignment to the "hair" was observed to align most closely with typical comic scenarios, where speech bubbles are often placed near a character's head for clarity. Based on this observation, we consistently used the "hair" prompt throughout our experiments and analyses in this paper. This uniform choice ensures consistency in bubble placement while leveraging the reliable performance of CLIPSeg in segmenting and assigning bubbles to relevant regions in a comic-style context.

\section{More qualitative evaluation}
In this section, we compared TaleDiffusion with StoryDiffusion \cite{zhou2024storydiffusion} and StoryGen \cite{liu2024intelligent} based on a 9-panel story (see \cref{fig:supp_comp}). It has been observed that StoryGen \cite{liu2024intelligent} neither follows the text prompt nor maintains character consistency. Also, it generates a lot of artifacts in each and every panel. Although StoryDiffusion \cite{zhou2024storydiffusion} follows the text prompt, it suffers from artifacts like a mutated hand and does not maintain the character as well as background consistency. In contrast, TaleDiffusion demonstrated superior performance across all evaluated aspects. The model not only produced a sequence of panels that were free from artifacts, but it also maintained consistent character appearances and backgrounds throughout the 9 panels. This level of consistency is crucial for creating a coherent and visually appealing comic story. Moreover, TaleDiffusion excels at dialogue rendering, enhancing the readability of the generated comic by accurately positioning and styling speech bubbles. This improvement in dialogue placement ensures that the flow of the story is clear and that the narrative is easy to follow for readers.

In summary, while both StoryGen and StoryDiffusion make significant contributions to story generation in the context of comics, TaleDiffusion outperforms both models by generating artifact-free, character-consistent, and visually coherent comics that effectively follow the narrative structure laid out by the text prompts. Its superior handling of dialogue rendering further elevates the readability and overall quality of the generated comic story.

\end{document}